\documentclass[journal]{IEEEtran}
\usepackage{cite}
\ifCLASSINFOpdf
  \usepackage[pdftex]{graphicx}
  \graphicspath{{figures/}}
\else
  \usepackage[dvips]{graphicx}
  \graphicspath{{figures/}}
\fi

\usepackage{amsmath}
\usepackage{array}
\usepackage{url}
\usepackage{booktabs}
\usepackage{array}
\newcolumntype{P}[1]{>{\centering\arraybackslash}p{#1}}
\usepackage{graphicx}
\usepackage{graphics}
\usepackage{amsfonts}
\usepackage{amssymb}
\usepackage{amsmath}
\usepackage{url}
\PassOptionsToPackage{hyphens}{url}
\usepackage{jmei_custom}
\usepackage{paralist}

\begin{document}
\title{EigenNetworks}
\author{Jonathan Mei
        and~Jos\'{e}~M.F.~Moura% <-this % stops a space
\thanks{This work partially funded by NSF grants CCF 1011903 and CCF
	1513936.}\\
\thanks{Department of Electrical and Computer Engineering,
Carnegie Mellon University,
Pittsburgh PA 15213 USA. The first author is now with Visby. email: jonathanmei@gmail.com, moura@ece.cmu.edu; Ph: (412)268-6341; fax: (412)268-3890.}}

\markboth{IEEE Transactions on Signal Processing}%
{Mei \MakeLowercase{\textit{et al.}}: EigenNetworks}


\maketitle

\begin{abstract}

	Many applications donot have the benefit of the laws of physics to derive succinct descriptive models for observed data. In alternative, interdependencies among $N$ time series $\{ x_{nk}, k>0 \}_{n=1}^{N}$ are nowadays often captured by a graph or network $G$ that in practice may be very large. The network itself may change over time as well (i.e., as $G_k$). Tracking brute force the changes of time varying networks presents major challenges, including the associated computational problems.  Further, a large set of networks may not lend itself to useful analysis. This paper approximates the time varying networks $\left\{G_k\right\}$ as weighted linear combinations of eigennetworks. The eigennetworks are fixed building blocks that are estimated by first learning the time series of graphs $G_k$ from the data $\{ x_{nk}, k>0 \}_{n=1}^{N}$, followed by a Principal Network Analysis procedure. The weights of the eigennetwork representation are eigenfeatures and the time varying networks $\left\{G_k\right\}$ describe a trajectory in eigennetwork space. These eigentrajectories should be smooth since the networks $G_k$ vary at a much slower rate than the data $x_{nk}$, except when structural network shifts occur reflecting potentially an abrupt change in the underlying application and sources of the data. Algorithms for learning the time series of graphs $\left\{G_k\right\}$, deriving the eigennetworks, eigenfeatures and eigentrajectories, and detecting changepoints are presented. Experiments on simulated data and with two real time series data (a voting record of the US senate and genetic expression data for the \textit{Drosophila
Melanogaster} as it goes through its life cycle) demonstrate the performance of the learning and provide interesting interpretations of the eigennetworks.
\end{abstract}
\label{ch:pna}
\section{Introduction}
\begin{quote}
``All entities move and nothing remains still.''\\
\raggedleft{--Heraclitus}
\end{quote}

In a world where change is the only constant, learning pairwise relationships among large ensembles of interacting agents from data they generate is often interesting but difficult. In many applications, data arises in contexts where physical laws are not available to derive useful models. An alternative paradigm is to capture relationships among data through sparse graph structures because they are interpretable by humans for advancing scientific understanding~\cite{deshpande_instantaneous_2011,sachs_causal_2005,valdes-sosa_estimating_2005}, as well as computationally beneficial for engineering efficient analytics algorithms~\cite{roweis_nonlinear_2000,livne_lean_2012,khan_neighborhood_2011,boykov_fast_2001}.

Many network time series models treat time series as stationary, where the parameters of the model--the networks--stay constant through time~\cite{bach_learning_2004,bolstad_causal_2011,dong_laplacian_2015,kalofolias_how_2016-1,segarramarquesmateosribeiro-2017,mei_signal_2017,mei_silvar:_2018,shen_kernel-based_2017}. We wish to extend some of these models to the non-stationary case in which these graph structures themselves vary in time. Time varying networks raise several challenges including the computational cost of tracking them, their succint representation, and extracting from them useful interpretations to close the loop with the application. To address these issues, we propose parsimonious representations of these time varying networks as weighted linear combinations of basis networks. We refer to the basis networks as principal networks or eigennetworks\footnote{\label{ftnote:footnote1}\textit{Eigenfaces} were introduced in \cite{sirovichkirby-1987} for face representation and identification and further developed in \cite{turkpentland-1991} for face detection and identification, see also \cite{belhumeurhespanhakriegman-1997} that explores eigenfaces and fisherfaces for face recognition and classification and \cite{chellappawilsonsirohey-1995} for a review of relevant work at the time.} and to the weights as eigenfeatures. The eigennetworks explain the variation of the graphs through a few basic component networks that are somehow fundamental to the underlying process that generates the time series. The time varying graphs describe a trajectory in eigenspace that is expected to be smooth since the graphs change at a much slower time scale than the data time series, except when rare but significant changes occur in the application or data sources. For example, an election where senate seats may flip from one party to another induces change in the voting pattern in the US senate, or the four stages of the \textit{Drosophila Melanogaster}\footnote{A vinegar fly, also commonly referred to as fruit fly.} metamorphosis should induce abrupt changes in its genetic expression data. In both applications, these events should be reflected in observable structural changes in the underlying networks. But in between these drastic events, we expect the structure of the networks to slowly vary, in the senate rolls, senators will mostly follow a party line position, and, in between life stages, the genetic expression data of the fruit fly exhibit small variations.  We consider these applications in Section~\ref{sec:exp}. The paper describes algorithms for
\begin{inparaenum}[1)]
\item learning the time series of graphs from the raw data;
\item how to extract the corresponding principal or eigennetworks, eigenfeatures and eigentrajectories from the estimates of the time varying networks; and
\item approaches to determine when significant changepoints occur.
\end{inparaenum}
 We illustrate and validate our concepts and methods with experiments on simulated data and with two real world applications, one involving the voting record of the US senate over a two year period  and the other the genetic expression data for the \textit{Drosophila Melanogaster} as it goes through its life cycle. These provide interesting opportunity to interpret eigennetworks in real life contexts.

%We also wish to capture significant changepoints of the graphs that otherwise vary smoothly between these points if they exist.
We explain now in more detail the approach.

\textbf{Eigennetworks and eigenfeatures.} Analysis of networks is usually difficult. The problem is compounded with varying networks. But the rate of change of the networks is expected to be slow relative to the original time series, it is then reasonable to believe that the collection of varying graphs can be well described through a small set of eigennetworks that are fundamental to the system. For example, in neuroscience, the functional connectivity of a subject in an experiment may present two distinct paradigms, when at rest and active, corresponding to the default mode and the task-positive networks~\cite{elton_task-positive_2015}; and in genetics, the gene regulatory network of an insect may present four main paradigms, corresponding to the developmental stages of egg, larva, pupa, and adult~\cite{davidson_regulatory_2006}. The description using eigennetworks allows a more compact summary of the process as well as a parsimonious representation, reducing the total parameter size of the final learning problem and providing more recognizable insights regarding the application data. Besides representation, the eigennetworks and the eigenfeatures can be used for other postprocessing network analysis, including classification, identification, or recognition, among others. Ideally, in certain applications, we may interpret each principal network individually as an eigenbehavior, shedding further light and helping to make sense of complex observed behaviors. Smooth variations of the networks are captured by time-varying weighted linear combinations of the eigennetworks.

Assuming that the time varying graphs have been learned from the time series, see next paragraph, we interpret each graph as a vector and desire a description of a vector subspace that efficiently captures most of the scatter variation in this set of graphs. The principal or  eigennetworks, like eigenfaces, see footnote~\ref{ftnote:footnote1} and~\cite{sirovichkirby-1987,turkpentland-1991,turk_face_1991}, span a lower dimensional subspace (``network space'') in which the networks describing the process live. There are myriad ways to find vectors spanning such subspaces~\cite{salakhutdinov_bayesian_2008,dueck_probabilistic_2004,hajinezhad_nonnegative_2016,candes_robust_2011}; we adopt Principal Component Analysis (PCA)~\cite{wold_principal_1987} as dimensionality reduction method that for a fixed number of basis components maximizes the scatter variation across the dimension reduced representations of the graphs. We apply a sparse version of PCA to networks and term this Principal Network Analysis (PNA).

\textbf{Estimating time varying networks and changepoints.}
Before applying PNA, we estimate the time varying networks. We present a generic nonlinear regression model to learn the graphs. In our framework, the graphs can be directed or undirected. We pose the learning as a regularized optimization problem incorporating desired structural properties such as sparsity of the graphs and low rank latent variables (accounting for trends or unmodeled effects on the observations). This is potentially a difficult and computationally expensive task, since we may need to learn a new network at every single time step! To make this vastly underdetermined problem tractable, additional assumptions are required. One common assumption takes the form that the model does not change ``too quickly'' (or ``too often'') through time. This can be characterized with respect to some metrics for signal smoothness~\cite{dahlhaus_locally_2012,chambolle_algorithm_2004}. This assumption tends to take on two primary variants: 1) bounded total variation~(TV); or 2) Lipschitz temporal gradient. The behavior of these variants can lead to strikingly different qualitative interpretations. Low total variation can be viewed as switching (along time) between networks at discrete (time) changepoints, whereas Lipschitz gradient can be viewed as a network slowly evolving in time.

%One method for enforcing smoothness on the learned model is to impose a hidden Markov model in a nonparametric Bayesian setting~\cite{willsky_nonparametric_2009}. This model is flexible in imposing low TV by allowing the number of switched networks to be learned simultaneously with the networks themselves. However, it requires Markov Chain Monte Carlo (MCMC) sampling to learn, which provides a more detailed picture about the final learned model than a point estimate but for which convergence of the sampler can in general be difficult to determine theoretically and in particular for high-dimensional regimes still tricky to ensure~\cite{andrieu_introduction_2003,rajaratnam_mcmc-based_2015}. Instead, we choose to
 For enforcing smoothness on the learned networks model, we  focus on optimization based approaches as the basis for our algorithms and analysis. There have been several attempts at extending optimization methods for estimating single networks to the non-stationary setting~\cite{song_time-varying_2009, kolar_estimating_2012}. However, these methods take a black and white approach to the smoothness assumption, with either totally smooth graphs or piecewise constant graphs. Our approach, based on eigennetworks and time-varying eigenfeatures (the weights of the linear combinations representing the networks) allows capturing both smoothness and determine abrupt changepoints. Clearly, simultaneously estimating both improves the overall performance of the approach, since, once a changepoint occurs, we will be able to better select the data to be used to estimate the network.

%Once the time varying graphs have been learned from the time series, we would like to estimate the eigennetworks. Thinking of each graph as a vector, we see that we desire a description of a vector subspace that efficiently captures most of the variation in this set of graphs. There are myriad ways to find vectors spanning such subspaces~\cite{salakhutdinov_bayesian_2008,dueck_probabilistic_2004,hajinezhad_nonnegative_2016,candes_robust_2011}; one of the most notable examples is Principal Component Analysis (PCA)~\cite{wold_principal_1987}. We apply a sparse version of PCA to networks and term this Principal Network Analysis (PNA).

\textbf{Summary.} We organize the paper as follows: section~\ref{sec:not} establishes notation and reviews background, in particular, from and extending the notation in~\cite{mei_silvar:_2018} from estimating a single network to estimating time varying networks, also reviewing Lipschitz and total variation based approaches to estimating time varying sparse networks; section~\ref{sec:smooth} introduces our framework for learning time-varying sparse graphs in the presence of latent effects; section~\ref{sec:changepoints} presents our local linear regression kernel based approach to determine changepoints; section~\ref{sec:pna} describes the estimation of the principal or eigennetworks and eigenfeatures from the time series of networks; section~\ref{sec:exp} details experimental results of our framework on simulated data and two sets of real data, one the voting record of the US senate in 2010-2011, before and after the 2010 November election, and the other genetic expression data for the \textit{Drosophila Melanogaster} that covers the four stages of its life cycle; and finally we conclude in section~\ref{sec:conc}.

\section{Notation and Related Works}
\label{sec:not}
In this section, we will describe the models that provide foundations for the methods we consider. Much of the notation follows from~\cite{mei_silvar:_2018}, which estimates a single network, so we provide a brief review. We will then introduce modifications to these models to capture non-stationarity and estimate a series of time-varying networks and compare this extension with several existing methods for time varying graphs in these settings. Finally, we outline the challenges that can arise in learning these graph structures.

\subsection{Generalized Linear Models}
\label{sec:bg:GLM}
The Generalized Linear Model (GLM) can be described using several parameterizations. We adopt the one based on the Bregman Divergence~\cite{bregman_relaxation_1967,banerjee_clustering_2005}. For observations $y_i\in\Rbb$ and $\x_i\in\Rbb^{p}$, let $\y=(y_1\ldots y_n)^\top$, $\X=(\x_1\ldots\x_n)$. The model is parameterized by 1) a non-linear link function $g=\nabla G$ for a convex function $G$; and 2) a vector $\a\in\Rbb^p$. We have the model written as
\begin{equation}
\label{eq:GLM_expec}
\Ebb[y_i | \x_i] = g \left(\a^\top \x_i \right),
\end{equation}
(note that some references use $g^{-1}$ as the link function where we use $g$).

For data $\{\x_i,y_i\}$ with conditionally independent $y_i$ given $\x_i$ (note that this is not necessarily assuming that $\x_i$ are independent), learning the model $\a$ assuming $g$ is known can be achieved via empirical risk minimization,
\begin{equation}
\label{eq:GLM_MLE}
\begin{aligned}
\what{\a} &= \argmin[\a] \sum\limits_{i=1}^{n}\;\left[ G_*\left(y_i\right) + G\left(\a^\top \x_i\right) - y_i \left(\a^\top \x_i\right) \right]
\end{aligned}
\end{equation}
where $G_*$ is the convex conjugate of $G$ (see~\cite{mei_silvar:_2018} for further details).

\subsection{Estimating Static Networks}
First, we extend the model to the multivariate case. Let $\y_i=\left(y_{1i} \; \ldots \; y_{mi}
\right)^\top$, $\g(\v)=\left(g_1(v_1) \; \ldots \; g_m(v_m) \right)^\top$, and the $m\times p$ matrix $\A=\left(\a_1 \; \ldots \; \a_m\right)^\top$. Consider the vectorization,
\begin{equation}
\label{eq:spars}
\begin{aligned}
\Ebb \left[y_{ji} | \x_i \right] &= g_j \left(\a_j^\top \x_i \right), \: j=1\cdots m \\
\Rightarrow \Ebb \left[\y_i | \x_i \right] &= \g \left(\A \x_i \right)\in\Rbb^m.
\end{aligned}
\end{equation}
In~\eqref{eq:spars}, the notation $\g \left(\A \x_i \right)$ stands for $\g \left(\A \x_i \right)=\left[g_1 \left(\a_1^\top \x_i \right)\cdots g_m \left(\a_m^\top \x_i \right)\right]^\top$. For the remainder of this paper, we make an assumption that all $g_j=g$ for notational simplicity, though the same analysis readily extends to the case where $g_j$ are distinct.

Now, consider an additional low rank matrix coefficient $\L$ in the following model,
\begin{equation}
\label{eq:lat_vars}
\begin{aligned}
\Ebb \left[\y_i | \x_i \right] &= \g \left((\A+\L) \x_i \right).
\end{aligned}
\end{equation}
The matrix $\L$ can be seen as incorporating the effects of some small number of latent variables $\z$ on the observed variables $\x$. That is, under a true model
\begin{equation}
\label{eq:lat_vars_1}
\begin{aligned}
\Ebb \left[\y_i | \x_i ,\z_i \right] &= \g \left(\A \x_i + \B\z_i \right),
\end{aligned}
\end{equation}
the coefficient matrix $\B$ induces $\L$ additive to $\A$ (the derivation is omitted due to space constraints; see~\cite{mei_silvar:_2018} for further details). For expositional clarity, we omit $\L$, noting that we can simply substitute $\A\leftarrow\A+\L$ if we desire such a model. However,in our experiments in section~\ref{sec:exp}, we work with the full model, with~$\L$ included.

We point out that formulating this problem as a generic regression allows us to consider graphs\footnote{In~\eqref{eq:spars} through~\eqref{eq:lat_vars_1}, we assume a generic model. When dealing with graphs, we will assume $m=N$ and $p=MN$, where~$N$ is the number of agents or nodes of the graph, $M$ is an integer, and the matrices~$\A$ and~$\L$ become of dimensions $N\times MN$.} as either directed or undirected. For example, to estimate directed graphs on a time series, we may use an autoregressive model of lag order $M$
\begin{align}
\label{eq:autoreg}
\Ebb\left[\x_k \big| \x_{k-1:k-M} \right] = \g\left( \left(\A_{k,1} \; \ldots \; \A_{k,M}\right) \x_{k-1:k-M} \right)
\end{align}
where $\x_{k-1:k-M}=(\x_{k-1}^\top \; \ldots \; \x_{k-M}^\top)^\top$.
To estimate undirected graphs, we may for example use an extemporaneous regression,
\begin{align}
\label{eq:extemp_reg}
\Ebb\left[x_{ki} \big| \x_{k\backslash	i} \right] = g\left( \left(\j_i\odot\a_{ki} \right) \x_k \right)
\end{align}
where $\j_i = (\1-\e_i)^\top$ and $\e_i$ is the $i$-th canonical basis vector, $\a_{ki}$ is the $i$-th row of $\A_k$. Viewed jointly for all indices $i$ of $\x_{k}$, in row $i$ we regress $\x_{k\backslash i}$ (all indices of $\x_k$ except for the $i$th) on $x_{ki}$ (the $i$th index of $\x_k$) by using $\J\odot\A_k$ (where $\j_i$ is the $i$-th row of $\J=\1\1^\top-\I$), enforcing contribution of 0's on the diagonal. This regression has a correspondence to the precision (inverse covariance) matrix. In general, we can also enforce any arbitrary symmetric nonzero structure $\J\in\{0,1\}^{N \times N}$ on $\A_k$ if we have corresponding prior structural information on the graph matrix. We summarize how these specific settings from~\eqref{eq:autoreg} and~\eqref{eq:extemp_reg} correspond to the generic notation used in equation~\eqref{eq:spars} in Table~\ref{tab:reg_subs}.

\begin{table}
	\begin{center}		
	\begin{tabular}{|c|c c|}
		\hline
		Notation in~\eqref{eq:spars} & AR (directed) in~\eqref{eq:autoreg} & Inv. Cov. (undirected) in~\eqref{eq:extemp_reg} \\	
		\hline
		$\y_k$ & $\x_k$ &  $\x_k$ \\	
		$\x_k$ & $(\x_{k-1}^\top \; \ldots \; \x_{k-M}^\top)^\top$ & $\x_k$ \\	
		$\A_k$ & $(\A_{k,1} \; \ldots \; \A_{k,M})$ & $\J\odot\A_k$ \\	
		\hline
	\end{tabular}
	\end{center}
	\caption{Summary of example regression settings and notation for estimating undirected and directed graphs}
	\label{tab:reg_subs}
\end{table}

When posing the learning as an optimization problem, the loss function from~\eqref{eq:GLM_MLE} naturally extends to the multivariate case. Other desired structural properties may be incorporated into the framework via regularization. Thus, we can pose the optimization as
\begin{equation}
\label{eq:multi_opt}
\what\A=\argmin[\A] \sum\limits_{j=1}^K f_j(\A) + \lambda h(\A)
\end{equation}
where
\begin{equation}
\label{eq:multi_loss}
f_j(\A) = \frac{1}{m} \left[ \1^\top \Big( \G_*(\y_{j}) + \G(\A\x_{j}) \Big) - \y_j^\top\A\x_{j} \right]
\end{equation}
Some examples for $h$ include symmetric sparsity for sparse partial correlations (conditional independencies) in a Gaussian graphical model~\cite{friedman_sparse_2008}, group sparsity corresponding to sparse Granger Causality~\cite{bolstad_causal_2011}, and commutativity corresponding to graph filters in the Discrete Signal Processing on Graphs framework~\cite{mei_signal_2017,sandryhaila_discrete_2013,sandryhaila_discrete_2014,sandryhaila_big_2014}. More detailed discussion of different choices for regularization and their assumptions will remain beyond the scope of this paper.

\subsection{Time-varying Networks}
To further reduce clutter when convenient, we define
\begin{equation}
\label{eq:Astack}
\begin{aligned}
\Acal &=(v(\A_{1}) \; \ldots \; v(\A_{K}))^\top \in \Rbb^{K \times mp},
\end{aligned}
\end{equation}
which collects matrices $\A_k\in\Rbb^{m \times p}$ through time index $K$, and $v(\A)$ denotes the vectorization of a matrix $\A$. Also, we use the various shorthand forms
\begin{equation}
\boldsymbol{\theta}_{kj} = \boldsymbol{\theta}_j(\A_k) = \boldsymbol{\theta}(\A_k, \x_{j})=\A_{k}\x_{j}
\end{equation}
so that the model is allowed to change at each time step $k$, and the interpretation for $\boldsymbol{\theta}_{kj}$ is the linear part of the model prediction made by using the network from time $k$ on data from time $j$. The actual model itself only makes sense at $k=j$, but we will see shortly why we wish to have the notational flexibility to describe these offset $k\ne j$ quantities.

Returning to our model, we have the familiar
\begin{equation}
\label{eq:TV_AR_model}
\Ebb[\y_k|\x_{k}] = \g\left( \A_{k}\x_{k} \right) = \g\left( \boldsymbol{\theta}_{kk} \right),
\end{equation}
with a possibly non-linear link function $g$ and an associated loss functional at time $k$ of the general offset form
\begin{equation}
\label{eq:offset_loss}
f_j(\A_k) = \frac{1}{m} \left[ \1^\top \Big( \G_*(\y_{j}) + \G(\boldsymbol{\theta}_{kj}) \Big) - \boldsymbol{\theta}_{kj}^\top\y_j \right]
\end{equation}
where $\G$ and $\G_*$ are vectorizations of $G$ and $G_*$ analogous to $\g$. As currently specified, the model is overdetermined. In the following, we review assumptions from prior art that restrict the model space and make the model estimation statistically tractable.

\subsection{Related Works}

Optimization based methods also come in several flavors. One approach to impose Lipschitz gradient is to simply use a kernel estimator~\cite{song_time-varying_2009} to obtain locally stationary solutions~\cite{dahlhaus_locally_2012} using our offset loss functionals from~\eqref{eq:offset_loss},
\begin{equation}
\what{\A}_k=\argmin[\A_k]\!\sum_{j=1}^{K}w_{kj} f_j(\A_k) + h(\A_k),
\end{equation}
where $w_{kj}$ is the kernel weight of a symmetric kernel with some bandwidth $\eta$ centered at $k$ evaluated at $j$, and $h(\A)$ is a regularizer on the sparsity of the network whose form we omit for simplicity. This is embarrassingly parallel, and allows estimating the network at a single time point without needing to compute those at other time points if deemed irrelevant. However, the smoothness of the kernel makes it more challenging to interpret for detecting discrete changepoints as the low TV setting allows.

Alternatively, an approach for enforcing low total variation~\cite{kolar_estimating_2012} is to regularize the difference between neighboring time points
\begin{equation}
\what{\Acal}=\argmin[\Acal]\!\sum_{k=1}^{K} f_k(\A_k) + \sum_{k=2}^K \| \A_{k}-\A_{k-1} \|_F .
\end{equation}
This group sparse regularizer encourages the difference between networks at adjacent time points to either be all zero or non-zero. This minimization is over all networks $\A_1,\ldots,\A_K$ through $\Acal$ defined in~\eqref{eq:Astack}. It couples the problem across time points, so that the solution is a joint estimator and is no longer so simply parallelized. Furthermore, while this model can capture significant changepoints, it is less able to describe smoother variation.

\subsection{Local Linear Regression}
\label{subsec:local_lin_reg}
The main conceptual workhorse for our approach is still kernel regression. However, we consider a locally linear regression instead of a vanilla kernel regression. This has the benefit of a lower bias near boundaries of the dataset, so it is natural to adopt for our purposes of changepoint detection. Changepoints can be thought of as a boundary that appears in the middle of the data; alternatively, the boundaries can be thought of as changepoints at the edges of the data domain.

The local linear regression estimator is defined similarly to the kernel estimator. For a clean signal $x_k$, noise signal $v_k$, and noisy observations $y_k=x_k+v_k$, the local linear regression estimator is
\begin{equation}
(\what{a}_k,\what{b}_k)=\argmin[a_k,b_k]\!\sum_{j=1}^{K}w_{kj} \|y_j-(a_k+(j-k)b_k)\|_2^2,
\end{equation}
where $\what{a}_k$ is an estimate for the value of the clean signal $x_k$ at time $k$, while $\what{b}_k$ represents an estimate for the slope of the clean signal $x_k$ at time $k$. This estimator, as compared to kernel regression, has a lower bias near the edges of the data~\cite{fan_local_1991}.

\subsection{Matrix Decompositions}

Consider the problem of expressing a given matrix as the product of two matrices
\begin{equation}
\Acal=\C\Bcal^\top
\end{equation}
where $\Acal\in\Rbb^{M\times N}$, $\C\in\Rbb^{M \times R}$, and $\Bcal\in\Rbb^{N \times R}$ for some $R>0$. In general, if $\textrm{rank}(\Acal)\le R$ for some positive integer $R$, then there can be infinite solutions for the matrices $\C$ and $\Bcal$. Particular structure may be imposed on $\C$ and/or $\Bcal$ to obtain desired properties or interpretations. These two matrices can be determined by any number of matrix factorization algorithms, such as a straightforward $R$-rank singular value decomposition in which
\begin{equation}
\begin{aligned}
&\Acal=\U\boldsymbol{\Sigma}\V^\top\\
\C=&\U\boldsymbol{\Sigma}^{\beta} \qquad \Bcal=\V\boldsymbol{\Sigma}^{1-\beta}
\end{aligned}
\end{equation}
for some choice of $\beta\in[0,1]$, typically.

This solution can be seen as a common starting point that can give rise to other solutions, since the existence of the SVD is guaranteed while its computation is stable~\cite{trefethen_numerical_1997}, and it yields one nice structure, with $\U$ and $\V$ defining orthonormal bases. However, since we expect in our application the networks to have different types of structure both temporally (smooth) and spatially (sparse), we would naturally expect any reasonably interpretable decomposition to exhibit qualitatively similar behavior. The SVD does not necessarily give us these behaviors. Thus, we consider matrix factorization techniques that can give rise to a temporally smooth $\U$ and a spatially sparse $\V$ (where we associate time with $\U$ and network structure with $\V$ indirectly via our arbitrary choice of how to stack $\Acal$ as in~\eqref{eq:Astack}).

There are many algorithms for performing matrix factorization, such as probabilistic frameworks that utilize sampling~\cite{salakhutdinov_bayesian_2008}. Again, we focus on optimization based frameworks. Many of these tend to center around variational methods~\cite{dueck_probabilistic_2004} or ADMM~\cite{boyd_distributed_2011, hajinezhad_nonnegative_2016}. Ultimately, we choose to use inertial Proximal Alternating Linear Minimization (iPALM)~\cite{pock_inertial_2016}. The iPALM method has similarities in form with ADMM, as they both utilize the separability of the objective function and computationally fast proximal operators. However, iPALM guarantees that certain types of nonconvex optimization problems, including regularized matrix factorization, converge to local minima via an adaptive choice of step size. We provide more details on implementation in section~\ref{sec:pna}.

\section{Smooth Graph Regression}
\label{sec:smooth}
We propose an optimization based approach that learns time varying graph structure jointly based on the full time series data. Our method offers the parallelizability of the previous smooth window-based approach.

\subsection{Formulation and computation}
\label{subsec:form_comp}

We formulate the locally linear regression based optimization problem,

\begin{equation}
\label{eq:sparse_smooth_changepoint}
\begin{aligned}
(\what{\Acal},\what{\Acal}') &=\underset{\Acal,\Acal}{\arg\!\min}\; F(\Acal,\Acal') + \lambda \textrm{h}(\Acal)
\end{aligned}
\end{equation}
for $i\in{\ell,c,r}$, where
\begin{equation}
\label{eq:sparse_smooth_changepoint_Fdef}
\begin{aligned}
F(\Acal,\Acal') \overset{\Delta}{=} \sum_{k=1}^K\sum_{j=1}^K w_{kj} f_j(\A_k \!+\!(j-k)\A'_k),
\end{aligned}
\end{equation}
and where $\textrm{h}(\Acal)$ is a regularizer on the structure of $\Acal$, the $N\times MN$ matrix $\A_k'$ corresponds to the instantaneous time derivative of $\Acal$ and is essentially a nuisance parameter in our setting. To reduce notational clutter, we have omitted the low-rank component, but the formulations follow exactly analogously from equation~\eqref{eq:lat_vars}, by considering the additive low-rank contribution from latent variables as $\Acal \!\leftarrow\! \Acal \!+\! \Lcal$ (again, see~\cite{mei_silvar:_2018} for details). For $\textrm{h}(\Acal)$, we can consider for example a group sparsity that corresponds to Granger causality if $\Acal$ is autoregressive (AR) coefficient matrices, similar to before,
\begin{equation}
\begin{aligned}
\textrm{h}(\Acal) &= \sum_{k=M+1}^{K}\sum_{i,j} \left\| \a_{kij} \right\|_2
\end{aligned}
\end{equation}
where $\a_{kij}=\left(\left[\A_{k,1}\right]_{ij} \; \ldots \; \left[\A_{k,M}\right]_{ij}\right)$, the vectors collecting the $ij$ elements of the respective AR coefficient matrices across all lag orders.

With this estimator, we can see that it still remains trivially parallelizable across time points $k$,
\begin{equation}
\label{eq:sparse_smooth_para}
\begin{aligned}
(\what{\A}_k,{{\what{\A}'}_k}) &=\underset{\A_k,
	\A'_k}{\arg\!\min}\; F_k(\A_k,\A'_k) + \lambda \textrm{h}_{1}(\A_k)
\end{aligned}
\end{equation}
where
\begin{equation}
\label{eq:sparse_smooth_para_Fdef}
\begin{aligned}
F_k(\A_k,\A'_k) \overset{\Delta}{=} \sum_{j=1}^K w_{kj} f_j(\A_k+(j-k)\A'_k)
\end{aligned}
\end{equation}
and where $\textrm{h}_1(\A_k)$ simply acts on one of the $\A_k$.

\subsection{Estimation Procedure}
Since our optimization problems~\eqref{eq:sparse_smooth_changepoint} are convex, to produce our estimates, we can use any number of convex algorithms to solve the formulation. For concreteness, we provide an outline for one such approach using proximal gradient methods.

The gradient computations required are, 	
\begin{equation}
\label{eq:partialF_kernel}
\begin{aligned}
&\frac{\partial F}{\partial \A_{k}} \overset{\Delta}{=} \E_{k} = \J\odot\sum_{j=1}^{K}w_{kj}\left(g(\boldsymbol{\theta}_{kj}) - \y_j\right)\x_{j}^\top \\
&\frac{\partial F}{\partial \A'_{k}} \overset{\Delta}{=} \H_{k} = \J\odot\sum_{j=1}^{K}(j-k)w_{kj} \left(g(\boldsymbol{\theta}_{kj}) - \y_j\right)\x_{j}^\top
\end{aligned}
\end{equation}
where $\J$ is the binary mask enforcing the 0 structure. We can let $\Ecal$ and $\Hcal$ collect and stack $\{ \E_{k} \}$ and $\{ \H_{k} \}$ (respectively) the same way as $\Acal$ collects and stacks $\{ \A_{k} \}$,
\begin{equation}
\begin{aligned}
&\Ecal=\left( v(\E_{1}) \; \ldots \; v(\E_{K}) \right)^\top\\
&\Hcal=\left( v(\H_{1}) \; \ldots \; v(\H_{K}) \right)^\top.
\end{aligned}
\end{equation}

Finally, we need the proximal operators for the regularizers,
\begin{equation}
\label{eq:proxh_kernel}
\begin{aligned}
&\textrm{prox}_{1}(\A_k, t) = \argmin[\Y] \frac{1}{2}\|\A_k-\Y\|_F^2 + t \textrm{h}_{1}(\Y)
\end{aligned}
\end{equation}

For $\textrm{h}_1(\A_k)=\sum_{i,j} \left\| \a_{kij} \right\|_2$, we have
\begin{equation}
\begin{aligned}
& \S_k = \textrm{prox}_{1}(\A_k, t)\\
\Longrightarrow& \s_{kij} = \begin{cases} \left( \frac{\left\| \a_{kij} \right\|_2 -t}{\left\| \a_{kij} \right\|_2\phantom{-t}} \right) \a_{kij}& \left\| \a_{kij} \right\|_2 > t\\
0& \left\| \a_{kij} \right\|_2 \le t
\end{cases},
\end{aligned}
\end{equation}
where the indexing of $\s_{kij}$ within $\S_k$ is the same as that of $\a_{kij}$ within $\A_k$, which corresponds to a form of group soft thresholding that can be performed quickly.

To put it all together,	algorithm~\ref{alg:TVG} describes the full proximal gradient implementation.
\begin{algorithm}[!h]
	\caption{TV Graphs using proximal gradient descent}
	\label{alg:TVG}
	\begin{algorithmic}[1]
		\Require Regularization parameter $\lambda>0$, max. iterations $t_{\max}$.
		\State Initialize $(\Acal,{\Acal'})=(\0,\0)$ or randomly, set iteration $t=0$, initial step size $s_0$.
		\While{not converged and $t<t_{\max}$} (in parallel over $k$)
		\State Compute gradients using equation~\eqref{eq:partialF_kernel}:
		\begin{equation}
		\begin{aligned}
		\E_k \leftarrow \frac{\partial F}{\partial \A_k} & \qquad \H_k \leftarrow \frac{\partial F}{\partial {\A'}_k}.
		\end{aligned}
		\end{equation}
		\State Compute proximal step
		\begin{equation}
		\begin{aligned}
		\A_k &\leftarrow \textrm{prox}_1(\A_k - s_t\E_k, s_t\lambda )\\
		\A'_k &\leftarrow \A_k' - s_t\H_k.
		\end{aligned}
		\end{equation}
		\State $t\leftarrow t+1$
		\EndWhile
		\State \Return $\what{\Acal}$
	\end{algorithmic}
\end{algorithm}

%\subsection{Changepoints}
\section{Changepoints}
\label{sec:changepoints}
In section~\ref{sec:smooth}, the estimation procedure handles smoothly varying networks. However, we may wish to  be able to capture abrupt structural changepoints. This will also have the advantage of limiting the estimate of the network from time series data from networks before or after they change significantly. We consider this here. We borrow a clever use of the local linear regression~\cite{qiu_jump-preserving_2003,gijbels_jump-preserving_2007} that estimates smooth 1D signals with discrete discontinuities. We can formulate several related problems with varying kernels,

\begin{equation}
\begin{aligned}
\left(\what{\A}^{(i)}_k, \what{\A}'^{(i)}_k \right) = \argmin[\A, \A'] \sum_{j=1}^K w^{(i)}_{kj} f_j\left(\A+(j-k)\A'\right)
\end{aligned}
\end{equation}
for $i\in\{\ell,c,r\}$ denoting left, center, and right, respectively, for which the difference between the estimators is in the kernel weights. Specifically, $w^{(c)}_{kj}$ is still a symmetric kernel centered at $k$ evaluated at $j$, but
\begin{equation}
\begin{aligned}
&w^{(r)}_{kj} = \begin{cases}
w^{(c)}_{kj} & j\ge k\\
0 & j < k
\end{cases}\\
&w^{(\ell)}_{kj}=w^{(r)}_{jk} .
\end{aligned}
\end{equation}

The intuition behind this suite of estimators is that each will perform best at different time points relative to the locations of changepoints. Consider a single changepoint at location $k'$. For some small enough $\kappa$ relative to the kernel bandwidth $\eta$, if $k\in[k'-\kappa, k')$, then both the right and center estimators will use data from two different networks from across the changepoint, while the left estimator will only use data from one network (to $0$th order); hence, we would expect the left estimator to be better. Similarly, we expect the right estimator to perform best when $k\in(k', k'+\kappa]$. Finally, with no changepoint we expect the center estimator to be the best since it uses more relevant data.

Given our intuition on change points, how do we quantitatively decide which estimator of the three to use? At each time point, we can compute the empirical residuals of each estimator for $i\in\{\ell,c,r\}$,
\begin{equation}
\label{eq:sparse_smooth_err}
\begin{aligned}
\what{\epsilon}_k^{(i)} \overset{\Delta}{=} \sum_{j=1}^K w^{(i)}_{kj} f_j\left(\what{\A}^{(i)}_k+(j-k)\what{\A}'^{(i)}_k\right)
\end{aligned}
\end{equation}

Letting $\gamma_c=\gamma\ge 0$ and $\gamma_\ell=\gamma_r=1$, we take the estimate
\begin{equation}
\label{eq:sparse_smooth_est}
\begin{aligned}
I_k &= \argmin[i\in \{\ell,c,r\}] \gamma_i\what{\epsilon}_k^{(i)}\\
\what{\A}_k &= \what{\A}_k^{(I_k)},
\end{aligned}
\end{equation}
where $\gamma$ theoretically should be chosen according to assumptions about problem settings, such as those on the noise variance, minimum magnitude of the changepoints, and kernel shape, to name a few. In practice, it could be chosen via some validation procedure. To elaborate, for $\gamma=0$, we always choose the central estimator and declare no changepoints (and may skip computing the $\ell$ and $r$ estimators), while for $\gamma \rightarrow \infty$ we never choose the central estimator (and may skip computing it). This range of $\gamma$ values draws out the ROC for the changepoint detector. In fact, it is shown in~\cite{qiu_jump-preserving_2003, gijbels_jump-preserving_2007} that, when $\lambda=0$, even $\gamma > 1$ implies (theoretically) never using the central estimator in the interior of the interval, and thus they advise against this.

Finally, the changepoints can be detected by considering the set of indices for which
\begin{equation}
\label{eq:sparse_smooth_chg}
\begin{aligned}
\what{\Jcal} = \left\{k: \left(I_k = \ell \right) \bigcap \left(I_{k+1} = r\right) \right\} .
\end{aligned}
\end{equation}

This tends to produce smooth behavior except near the changepoints, where the one-sided estimators are used. As an extra post-processing step, we may compute the central estimate on the boundaries of the contiguous segments to ensure smoothness in these regions.

This intuition leads to a speedup of the original detection scheme for generic $\gamma$ as well. We can first find all the potential changepoints as determined by these crossovers. Then \textit{only} at the two time points on either side of the crossover, compute the central estimators to verify that the left and right estimators are indeed better fit than the central (i.e., that $I_k\ne 0$ for either of these points).

This leads to a modified algorithm,
To put it all together,	algorithm~\ref{alg:TVG} describes the full proximal gradient implementation.
\begin{algorithm}[!h]
	\caption{TV Graphs with changepoints}
	\label{alg:TVG_jump}
	\begin{algorithmic}[1]
		\Require Threshold factor $\gamma>0$, regularization parameter $\lambda>0$, max. iterations $t_{\max}$.
		\State Initialize $(\Acal^{(i)},{\Acal'}^{(i)})=(\0,\0)$ or randomly, set iteration $t=0$, initial step size $s_0$.
		\While{not converged and $t<t_{\max}$} (in parallel over $k$)
		\State Compute gradients using equation~\eqref{eq:partialF_kernel}:
		\begin{equation}
		\begin{aligned}
		\E_k^{(i)} \leftarrow \frac{\partial F^{(i)}}{\partial \A_k^{(i)}} & \qquad \H_k^{(i)} \leftarrow \frac{\partial F^{(i)}}{\partial {\A'}_k^{(i)}}.
		\end{aligned}
		\end{equation}
		\State Compute proximal step
		\begin{equation}
		\begin{aligned}
		\A_k^{(i)} &\leftarrow \textrm{prox}_1(\A_k^{(i)} - s_t\E_k^{(i)}, s_t\lambda )\\
		{\A'_k}^{(i)} &\leftarrow {\A'_k}^{(i)} - s_t\H_k^{(i)}.
		\end{aligned}
		\end{equation}
		\State $t\leftarrow t+1$
		\EndWhile
		\For{$k\in\{M+1, \ldots, K\}$} (in parallel)
		\State Compute empirical errors and select indices
		\begin{equation}
		\begin{aligned}
		\wtil{\epsilon}_k^{(i)}& \!\!=\! \sum_{j=1}^K w^{(i)}_{kj} f_j\left(\what{\A}^{(i)}_k+(j-k)\what{\A}'^{(i)}_k\right)\\
		I_k &= \argmin[i\in \{\ell,c,r\}] \gamma_i\wtil{\epsilon}_k^{(i)}\\
		\what{\A}_k &= \what{\A}_k^{(I_k)},
		\end{aligned}
		\end{equation}
		\EndFor
		\State Compute changepoints
		\begin{equation}
		\begin{aligned}
		\what{\Jcal} = \left\{k: \left(I_k = \ell \right) \bigcap \left(I_{k+1} = r\right) \right\} .
		\end{aligned}
		\end{equation}
		\State (Optional for smoothness) Compute central estimate near boundaries
		\State \Return $\what{\Acal}, \what{\Jcal}$
	\end{algorithmic}
\end{algorithm}

\section{Principal Network Analysis}
\label{sec:pna}
While our method for estimating time-varying networks is fairly general, in many cases the graphs vary in time. Network analysis is in general difficult, but analysis of such ensemble of networks is even more difficult and may obscure interpretation of the application. We discuss an alternative simpler reprsentation of the time varying networks as a (time varying) weighted linear combination of some small set of so called ``Principal Networks'' or eigennetworks.
Letting
\begin{equation}
\label{eq:eigencomponents}
\begin{aligned}
\c_{k}&=(c_{k}^{(1)} \; \ldots c_{k}^{(r)} \;\ldots c_{k}^{(R)})^\top \in \Rbb^{R \times 1}\\
\C &=\left(\c_{1} \; \ldots \; \c_{K}\right)^\top \in \Rbb^{K \times R}\\
\B^{(r)}& \in\Rbb^{m\times p} \\
\Bcal\! &=\!\!\left(\!v\left(\B^{(1)}\right) \! \ldots \! v\left(\B^{(r)}\right)\! \ldots \; v\left(\B^{(R)}\right)\!\right) \!\! \in\!\! \Rbb^{m p \times R},
\end{aligned}
\end{equation}
we can decompose a set of time varying graphs  ($k$th row of $\Acal$ is the vectorization of network~$\A_k$) as
\begin{equation}
\label{eq:eigennetworks-1}
\Acal=\C\Bcal^\top.
\end{equation}
Row~$\c_{k}^\top$ of~$\C$ is the vector of~$R$ weights or eigenfeatures at time~$k$ for network~$\A_k$. Column $\c^{(r)}$ of~$\C$ is the time variation of the~$r$th eigenfeature $c_{k}^{(r)}$. The~$R$ rows $v\left(\B^{(r)}\right)^\top$ of~$\Bcal$ are the~$R$ vectorized principal or eigennetworks. Equation~\eqref{eq:eigennetworks-1} expresses each network~$\A_k$ ($k$th row of~$\Acal$) as a weighted linear combination of the~$R$ eigennetworks (rows of~$\Bcal^\top$). The matrices $\C$ and $\Bcal$ can be determined by any number of matrix factorization algorithms. We use an inertial version of Proximal Alternating Linearized Minimization (PALM, or iPALM for the inertial version) because of its theoretical guarantee of convergence to stationary points without requiring a tuning parameter~\cite{bolte_proximal_2014,pock_inertial_2016}.

We perform matrix factorization on the output from algorithm~\ref{alg:TVG} directly. Before we pose the optimization problem, we introduce the regularization to be used. For some $R>0$, let
\begin{equation}
\widetilde{\A}_{k}=\widetilde{\A}(\Bcal,\c_k)=\sum_{r=1}^{R} c_{k}^{(r)}\B^{(r)} .
\end{equation}

Let our sparsity regularizer be $h_1(\Bcal)$. For concreteness, in the autoregressive (AR) setting we may choose a group regularization corresponding to Granger Causality,
\begin{equation}
\begin{aligned}
h_1(\Bcal) &= \sum_{r=1}^{R} \left\| \b_{rij} \right\|_2
\end{aligned}
\end{equation}
where similarly to before $\b_{rij}=\left(\left[\B^{(r)}_{1}\right]_{ij} \; \ldots \; \left[\B^{(r)}_{M}\right]_{ij}\right)$, the vectors collecting the $ij$ elements of the respective AR coefficient matrices across all lag orders.

Also, let our low rank regularizer be
\begin{equation}
\begin{aligned}
h_*(\widetilde{\Acal}) &= \frac{1}{2} \left( \left\| \C \right\|_F^2 + \left\|\Bcal\right\|_F^2 \right).
\end{aligned}
\end{equation}

Now, we give the matrix factorization optimization problem,	
\begin{equation}
\label{eq:mf}
\underset{\Bcal,\C}{\arg\!\min} F_{\textrm{mf}}(\Bcal,\C)\!\! \overset{\Delta}{=} \!\!\frac{1}{2}\|\what{\Acal}\!-\!\C\Bcal^\top\|_F^2\! +\! \lambda_{*} h_*(\widetilde{\Acal})\! + \!\lambda_{1} h_{1}(\Bcal),
\end{equation}
where the subindex mf stands for matrix factorization. The Principal Network Analysis (PNA) yields sparse $\Bcal$ that can be viewed as our principal networks, which fundamentally underlie the process and are present at any given time point in some weighted linear combination with each other, with temporally smoothly varying weights or eigenfeatures $\C$ potentially with changepoints as a result of $\Acal$ being estimated in the same way. We could alternatively regularize $\C$ to be smooth (e.g., via a grouped sparse version of Total Generalized Variation~\cite{bredies_total_2010}) while not regularizing $\Bcal$ or even to regularize both. However, since we perform this factorization on the estimated $\what\Acal$, which is already sparse and smooth, it is somewhat redundant and only increases the computational complexity.

To solve our nonconvex problem~\eqref{eq:mf} using iPALM, we require a gradient computation on the smooth parts of the loss function and proximal operator computations for the non-smooth parts. Let
\begin{equation}
\label{eq:defH}
H(\Bcal,\C)=\frac{1}{2}\|\what{\Acal}-\C\Bcal^\top\|_F^2 + \lambda_{*} h_*(\widetilde{\Acal}).
\end{equation}

The gradient computations required are	
\begin{equation}
\label{eq:partialH}
\begin{aligned}
&\frac{\partial H}{\partial \Bcal} = \Qcal^\top\C + \lambda_*\Bcal\\
&\frac{\partial H}{\partial \C} = \Qcal\Bcal + \lambda_*\C ,
\end{aligned}
\end{equation}
where $\Qcal=\C\Bcal^\top - \what{\Acal}$. To apply iPALM, we also need to compute an upper bound for the following Lipschitz constants,	
\begin{equation}
\begin{aligned}
\label{eq:Lipschitz_Def}
&L_\Bcal(\C) \overset{\Delta}{=} \frac{\left\|\frac{\partial H}{\partial \Bcal}(\Ucal,\C)\!-\!\frac{\partial H}{\partial \Bcal}(\Vcal,\C)\right\|_F}{\left\|\Ucal-\Vcal\right\|_F}  \\
&L_\c(\Bcal) \overset{\Delta}{=} \frac{\left\|\frac{\partial H}{\partial \C}(\Bcal,\U)\!-\!\frac{\partial H}{\partial \C}(\Bcal,\V)\right\|_F}{\left\| \U-\V \right\|_2} .
\end{aligned}
\end{equation}

Thus we consider
{\small\begin{align}
\label{eq:Lipschitz_cB}
\left\|\frac{\partial H}{\partial \Bcal}(\Ucal,\C)\!-\!\frac{\partial H}{\partial \Bcal}(\Vcal,\C)\right\|_F \!\! &\!=\! \|(\Ucal \!-\! \Vcal)\C^\top \! \C \!+\! \lambda_*(\Ucal \!-\! \Vcal)\|_F \nonumber\\
&\le \left(\|\C^\top\C\|_F \!+\! \lambda_* \right)\|\Ucal \!-\!\Vcal\|_F \nonumber\\
\Longrightarrow L_\Bcal(\C) &\le \overline{L}_\Bcal(\C) \overset{\Delta}{=} \|\C\|_F^2 + \lambda_* .
\end{align}
}Similarly,
{\small\begin{align}
\label{eq:Lipschitz_Bc}
\left\|\frac{\partial F}{\partial \C}(\Bcal,\U)\!-\!\frac{\partial F}{\partial \C}(\Bcal,\V)\right\|_F \!\! &\!=\! \|(\U \!-\! \V)\Bcal^\top\Bcal \!+\! \lambda_* (\U \!-\! \V) \|_F \nonumber\\
\Longrightarrow  L_\C(\Bcal) &\le \overline{L}_\C(\Bcal) \overset{\Delta}{=} \left\|\Bcal\right\|_F^2 + \lambda_* .
\end{align}}

These upper bounds on the Lipschitz constants adaptively determine the step sizes in each coordinate block at each iteration, and thus they are slightly loose but more straightforward to compute than some other possible tighter bounds. Thus iPALM requires additional set-up up front to apply, but does not need tuning for step sizes, as compared to ADMM, which does not require the Lipschitz computation, but has a step size or learning rate parameter that can require properly tuning or setting in more challenging non-convex problems~\cite{boyd_distributed_2011,hajinezhad_nonnegative_2016}.

Finally, we need the proximal operators for the regularizers,
\begin{equation}
\label{eq:proxh}
\begin{aligned}
&\textrm{prox}_{1}(\Bcal, t) = \argmin[\Ycal] \frac{1}{2}\|\Bcal-\Ycal\|_F^2 + t \textrm{h}_{1}(\Ycal)\\
&\textrm{prox}_{s}(\C, t) = \C ,
\end{aligned}
\end{equation}
which can be solved quickly using soft thresholding. Putting everything together, we have algorithm~\ref{alg:PNA}.
\begin{algorithm}
	\caption{PNA using iPALM with two coordinate blocks}
	\label{alg:PNA}
	\begin{algorithmic}[1]
		\State Let $t=-1$. Set $\epsilon=\infty$, and tolerance $\delta>0$. Overloading superscripts to denote the iteration, initialize $\C^{-1}=\C^{0}=\textrm{rand}(K \times R)$ and $\Bcal^{-1}=\Bcal^{0}=\textrm{rand}(R \times mp)$ or with R-rank SVD.
		\While{$\epsilon \ge \delta$ and $t<t_{\max}$}
		\State $t\leftarrow t+1$
		\State Set inertial coefficient $\zeta = \frac{t}{t+3}$
		\State \vspace{-1.375em}~\begin{flalign*}
		\quad\;\begin{aligned}
		&\textrm{Update }\Bcal\textrm{ with fixed }\C\textrm{:}\\
		&\quad\textrm{Use inertia: } \Ycal = \Bcal^{t} + \zeta (\Bcal^{t}-\Bcal^{t-1})\\
		&\quad\textrm{Compute gradient and Lipschitz constant for}\\
		&\quad\Bcal\textrm{ from~\eqref{eq:partialH} and~\eqref{eq:Lipschitz_Bc}:}\\
		&\quad\qquad\Gcal_b=\partial_\Bcal{F}(\Ycal) \qquad\qquad\overline{L}_b=\overline{L}_\Bcal(\C^{t})\\
		&\quad\textrm{Apply Proximal operator from~\eqref{eq:proxh}:}\\
		&\quad\qquad \Bcal^{t+1} = \textrm{prox}_{1}(\Ycal - \Gcal_b/\overline{L}_b, \lambda_{1,b}/\overline{L}_b)
		\end{aligned} &&
		\end{flalign*}
		\State \vspace{-1.375em}~\begin{flalign*}
		\quad\;\begin{aligned}
		&\textrm{Update }\C\textrm{ with fixed }\Bcal\textrm{:}\\
		&\quad\textrm{Use inertia: } \Z = \C^{t} + \zeta (\C^{t}-\C^{t-1})\\
		&\quad\textrm{Compute gradient and Lipschitz constant for}\\
		&\quad\c\textrm{ from~\eqref{eq:partialH} and~\eqref{eq:Lipschitz_cB}:}\\
		&\quad\qquad\G_c=\partial_\C{F}(\z) \qquad\qquad\overline{L}_c=\overline{L}_\C(\Bcal^{t+1})\\
		&\quad\textrm{Apply Proximal operator from~\eqref{eq:proxh}:}\\
		&\quad\qquad\C^{t+1} \!=\! \textrm{prox}_{s}(\Z - \G_c/\overline{L}_c, \lambda_{s,c}/\overline{L}_c)\\
		&\quad\qquad\phantom{\C^{t+1}} = \Z - \G_c/\overline{L}_c
		\end{aligned} &&
		\end{flalign*}
		\State Compute error
		\begin{align*}
		\epsilon \leftarrow {\scriptsize\left\|\left(\begin{aligned}\Bcal^{t+1} \\ \C^{t+1}\end{aligned}\right)-\left(\begin{aligned}\Bcal^{t} \\ \C^{t}\end{aligned}\right) \right\|_F}\left/{\scriptsize\left\|\left(\begin{aligned}\Bcal^{t} \\ \C^{t}\end{aligned}\right)\right\|_F}\right.
		\end{align*}
		\EndWhile\\
		\Return $(\Bcal^{t}, \C^{t})$
	\end{algorithmic}
\end{algorithm}			

We note that a natural direction for an interesting line of future work would be to combine these two steps of first estimating a time varying graph and then performing matrix factorization into a single joint formulation that directly learns the factors, or iterates between the two steps as subproblems. This could bridge the gap between a full temporally coupled solution and its computation, since the factorized form would have fewer total parameters to estimate.

\section{Experiments}
\label{sec:exp}
We test the time varying graph estimation on simulated and two real data sets. We perform principal network analysis on the results of a US Senate voting record data set to yield the principal networks that we show have interpretable meaning in the context of the contemporary political climate. We also analyze gene interaction networks for the common fruit fly, the \textit{Drosophila  Melanogaster}, as it goes through the four metamorphosis stages of its life.

\subsection{Simulated Data}
%We generated networks of size $N=25$ of length $K=250$ and data from an autoregressive process of order $M=2$. We generated $\A\in\Rbb^{MN\times N}$ randomly as Erd\"{o}s-R\'{e}nyi in structure and $\Ncal(0,1)$ in weight with a diagonal uniformly generated in $[1,2)$. We generate the time-varying weights or eigenfeatures as piecewise constants with magnitude generated uniformly at random in $[2,4)$ plus a sinusoid of magnitude also generated uniformly at random in $[1/4,1/2)$, in which changepoints are generated uniformly at random in $\left[\frac{K-M}{S+1}(i-\frac{1}{8}),\frac{K-M}{S+1}(i+\frac{1}{8})\right)$ for $i=1,\ldots,S$ where $S$ is the number of changepoints.
%\input{simulatedexp}

 We explain how we generate the time varying networks $\left\{\A_k\right\}$ and the data $\left\{\x_k\right\}$, $k=1,\cdots, K$. For lack of space, we report results for the following choice of parameters: number of nodes $N=25$; number of eigennetworks $R=2$; number of change points $S=1$; order of the autoregressive $M=2$; time window $K=250$. From equation~\eqref{eq:autoreg}, $\A_k=\left(\A_{k,1}\, \A_{k,2}\right)$. At each~$k$, $\A_k$ will be a weighted linear combination of~$R=2$ eigennetworks. From equation~\eqref{eq:eigencomponents}, at each~$k$, we need the eigenfeature or weight $\c_{k}=\left(c_{k}^{(1)} \, c_{k}^{(2)}\right)^\top \in \Rbb^{2 \times 1}$ and the two eigennetworks $\B^{(r)} \in\Rbb^{2\times 25}$, $r=1,2$.
 
 We generate the $R=2$ eigennetworks $\B^{(r)}\in\Rbb^{25\times 50}$ as weighted Erd\"{o}s-R\'{e}nyi networks with probability of an edge $.025$ and asymetric random edge weights drawn from a normal distribution $\Ncal(0,1)$  with diagonal entries uniformly generated in $[1,2)$. In this study, we keep the eigennetworks fixed across the time window $k=1,\cdots,250$. We then generate changepoints  uniformly at random in $\left[\frac{K-M}{S+1}(i-\frac{1}{8}),\frac{K-M}{S+1}(i+\frac{1}{8})\right)$ for $i=1,\ldots,S$ where $S$ is the number of changepoints (in this study $S=1$). At each time step~$k$, the eigenfeatures or weights $\c_{k}=\left(c_{k}^{(1)} \, c_{k}^{(2)}\right)^\top$ of the linear combination are the sum of a fixed level and a sinusoidal term. Because we have two eigennetworks, we have two weights at each~$k$. The fixed level of each weight $c_{k}^{(j)}$, $j=1,2$, and in between changepoints, is kept constant and is drawn uniformly at random from $[2\,\,4)$. The parameters of each of the two sinusoids do not change in the whole time window, i.e., they remain the same across the changepoint. The amplitude of each sinusoid is drawn uniformly from the interval $[1/4\,\,1/2)$, the period being the same for both and equal to~$250$, the duration of the time window. The phase of one of the sinusoids is taken to be zero and that of the other to be $\pi/4$. By linear combination of the two (fixed)  Erd\"{o}s-R\'{e}nyi $\B^{(r)}$ eigennetworks weighted by the $K=250$ eigenfeatures $\c_{k}=\left(c_{k}^{(1)} \, c_{k}^{(2)}\right)^\top$ so generated, we created $K=250$ time varying networks $\A_k=\left(\A_{k,1}\, \A_{k,2}\right)$, each of size $N=25\times50=MN$. These time varying networks $\left\{\A_k\right\}$ are all created by the same structure (fixed eigennetowrks), but vary continuously because of the sinusoidal terms in the features $\c_{k}=\left(c_{k}^{(1)} \, c_{k}^{(2)}\right)^\top$, except at the changepoint when the weights jump as the magnitude levels and the amplitudes of the sinusoids jump. This makes it a fairly difficult setting conceptually, as the structure of the eigennetworks remains constant. We also generated data supported on these networks according to an autoregressive process of order $M=2$. 

\begin{figure}[thb]
	\includegraphics[width=0.95\columnwidth]{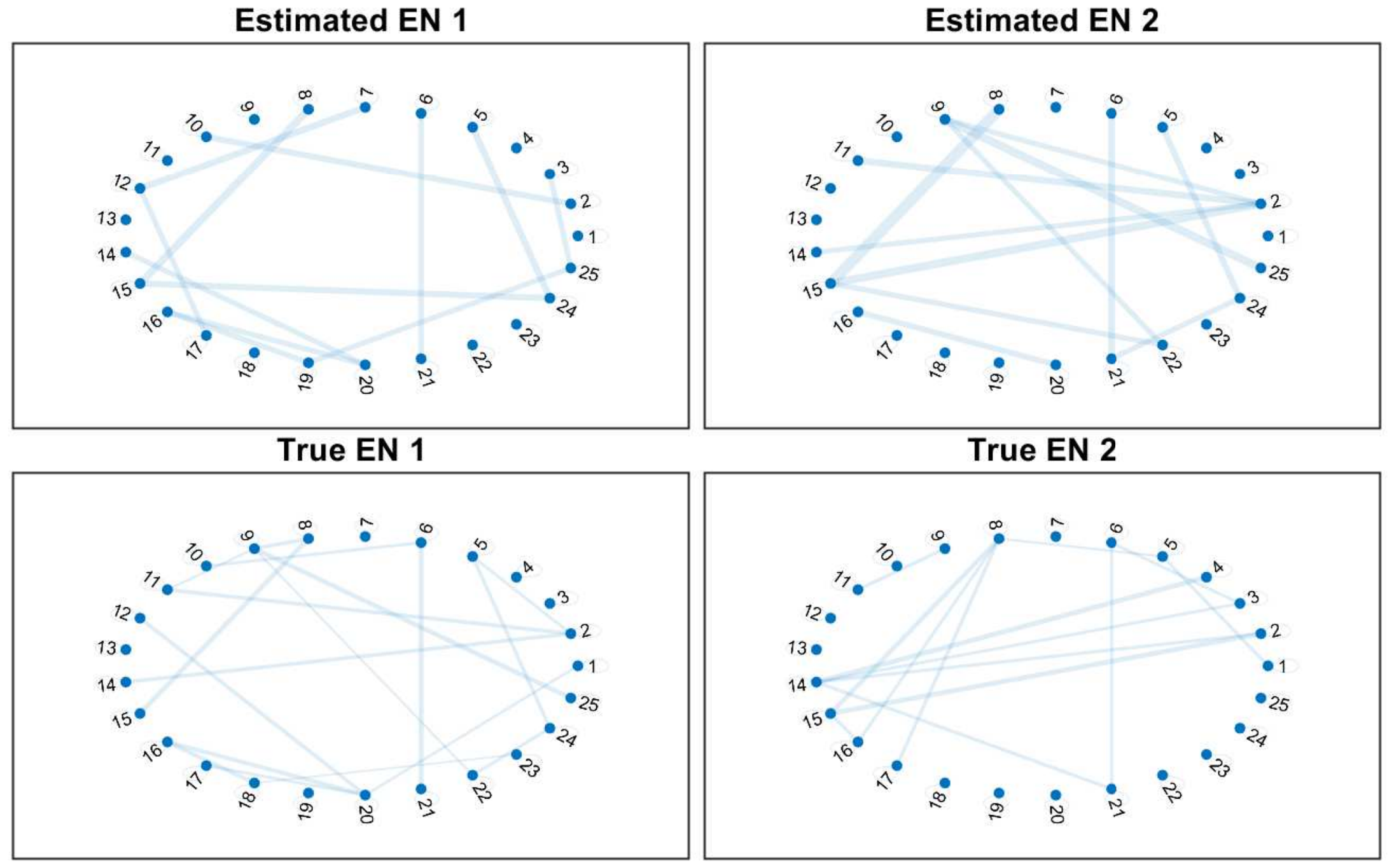}
	\caption{Graphs of the top 25 edges of the estimated and true eigennetworks}
	\label{fig:pna:toy_true_vs_est}
\end{figure}
Figure~\ref{fig:pna:toy_true_vs_est} shows the true and estimated eigennetworks used to generate the data. Structure-wise, the eigennetworks are estimated fairly well; the first eigennetwork has a probability of false alarm\footnote{Probability of false alarm of an edge is the probability that an estimated edge is not present in the true eigennetwork. Likewise, probability of detection is the probability that an edge in the true eigennetwork is present in the estimated eigennetwork. Because edges are weighted, we threshold the weights,}  $P_{\scriptsize\textrm{FA}}=0.029$ and a probability of detection $P_{\scriptsize\textrm{D}}=0.73$, while the second has a $P_{\scriptsize\textrm{FA}}=0.036$ and $P_{\scriptsize\textrm{D}}=0.72$. For visualization, we desire low false alarm to avoid clutter and end up at an operating point with conservative detection. We also note that this is a difficult setting since both networks are strongly active in the first half of the time series while less active in the second half. Thus the two networks are difficult to distinguish in noise, and we see some edges present in both estimates that are only present in one true eigennetwork (e.g., between nodes 16 and 20). A fuller characterization of the ROC would be an interesting direction for future work.

\begin{figure}[thb]
	\includegraphics[width=0.95\columnwidth]{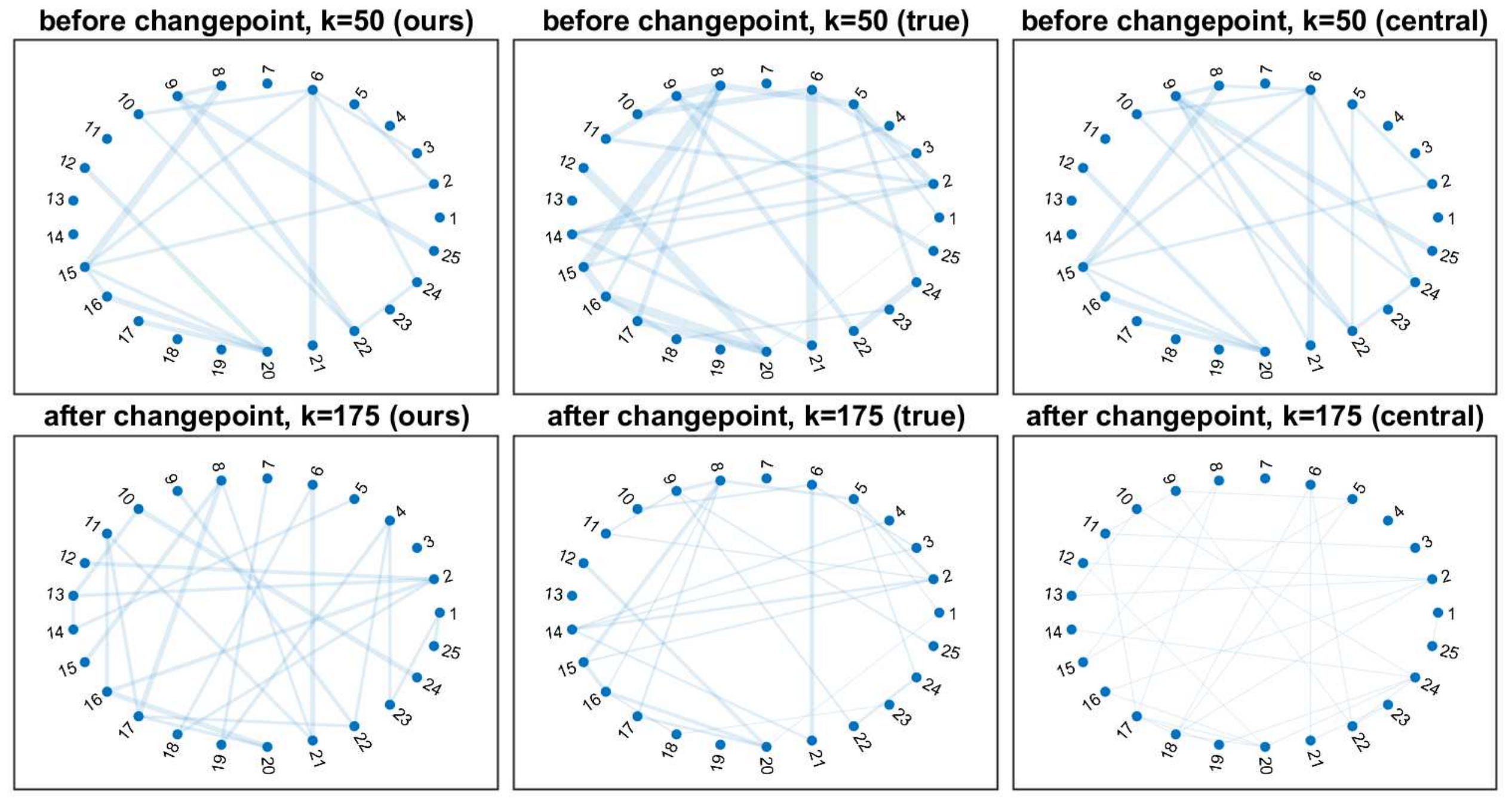}
	\caption{Graphs of the top 50 edges}
	\label{fig:pna:toy_graphs}
\end{figure}
\begin{figure}[thb]
	%\hspace{-0.075\columnwidth}
	\includegraphics[width=1.1\columnwidth]{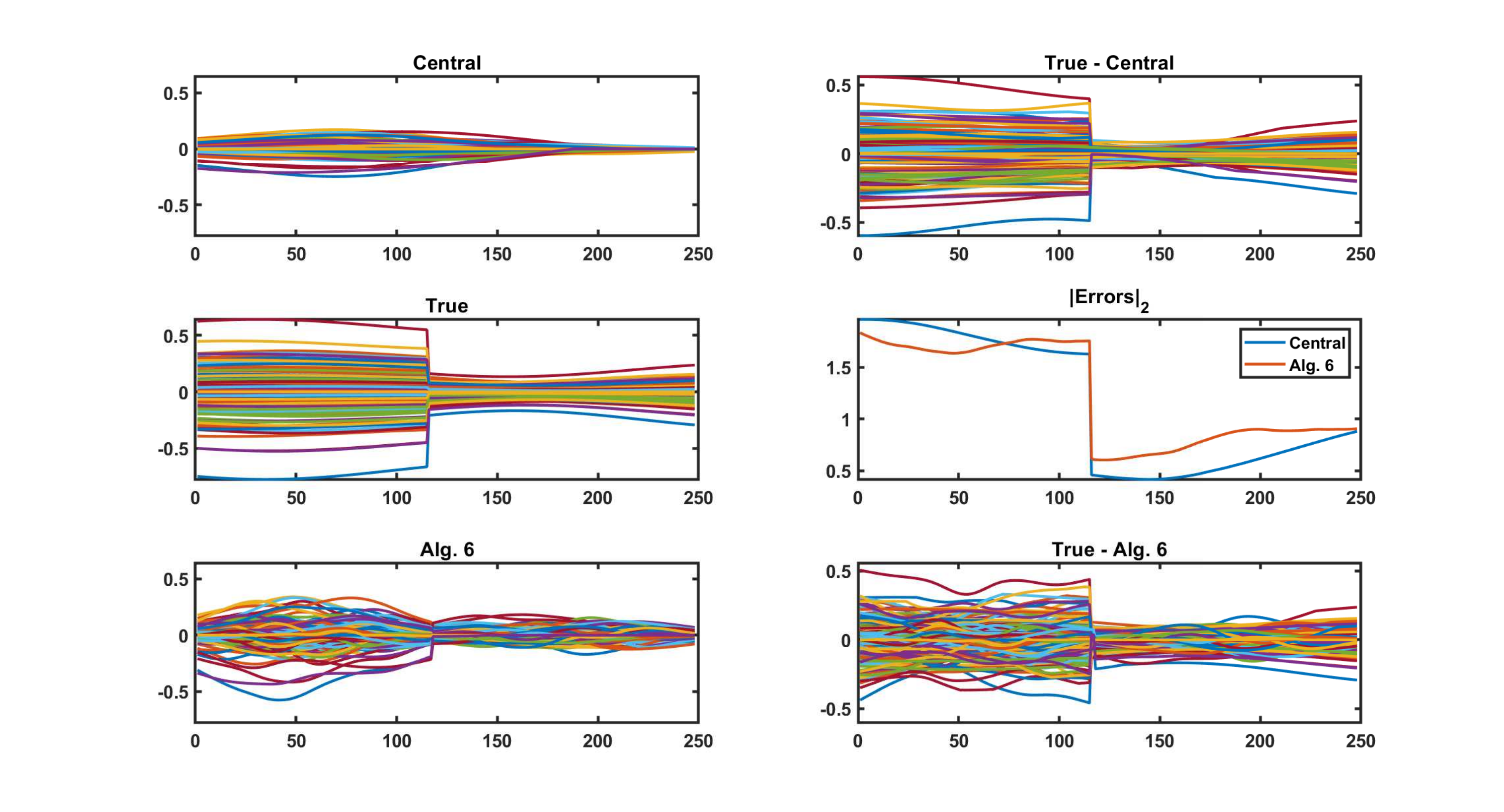}
	\caption{Time series of elements of graph adjacency matrix and corresponding errors}
	\label{fig:pna:toy_plots}		
\end{figure}

Figure~\ref{fig:pna:toy_graphs} shows that near the detected changepoint, using the same sparsity regularization parameter $\lambda$, the networks estimated with the changepoints are better able to capture the weights on either side of the changepoint, while the central estimator seems to average too strongly across the changepoint, resulting in weaker edges. We show in figure~\ref{fig:pna:toy_plots} the performance of our estimation method as compared to the central estimator for $\lambda=0.1$ and $w^{(c)}_{kj}\propto e^{-\frac{(j-k)^2}{200}}$.
In figure~\ref{fig:pna:toy_plots} each line represents the time series of the weights corresponding to one edge. We see that using our method, the changepoint is estimated well, with the overall error comparing favorably to that of the normal central kernel estimate. We note that the estimated magnitudes of the edge weights for the graph are shrunken as a result of the regularization, and there is some bias towards $0$ at the boundaries. This suggests an extension of the method would be to let the regularization parameter $\lambda$ vary through the interval rather than stay constant. A proper time profile of the $\lambda$ value would be of interest.
\subsection{US Senate Voting Data}
\begin{figure}[!h]
	\includegraphics[width=0.95\columnwidth]{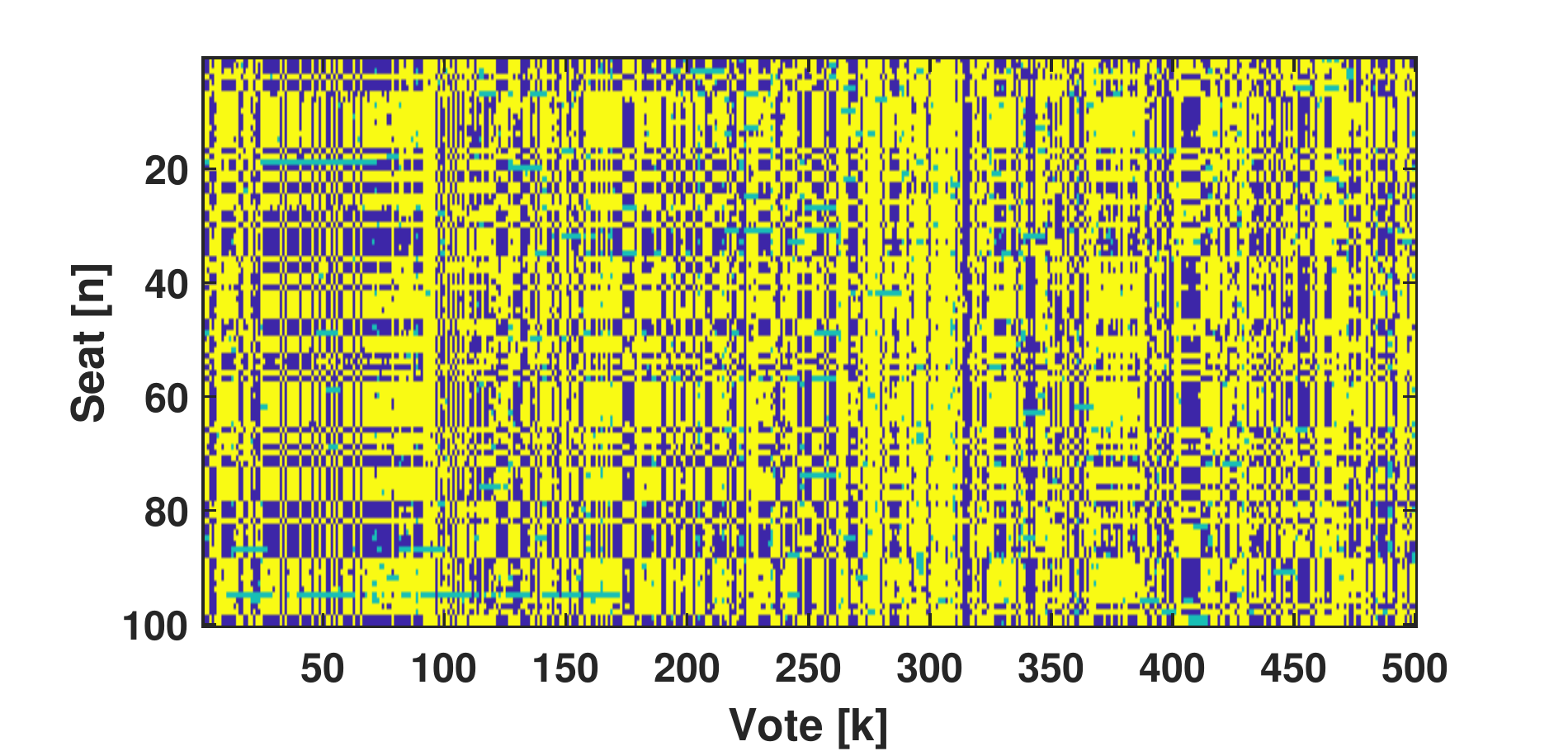}
	\caption{Time series of votes by seat, yellow (+1) for Yea, blue (-1) for Nay, and green (0) for abstention/vacant seat.}
	\label{fig:votes} 	
\end{figure}

We also compiled $K=500$ votes from the United States senate roll call records of sessions 111-112, corresponding to a period from 2010-2011~\cite{noauthor_u.s._2010,noauthor_u.s._2011}. There are $2$ senators from each of $50$ states, so that there are at most $N=100$ senators voting on each item. However, instead of directly tracking a growing and shrinking network composed of individual senators, we track the seat that the senator occupies, so that the time series of votes generated by senators continue being produced by their replacements. We treat yes votes as $+1$, no votes as $-1$, and abstentions/vacancies as $0$ so that the time series is $\X\in\{-1,0,+1\}^{100 \times 500}$ (see figure~\ref{fig:votes}). 

{\small
	\begin{table}[!h]
		\centering
		\caption{Examples of unanimous/procedural votes}
		\label{tab:votes}
		\begin{tabular}{@{}P{0.5in}P{0.5in}p{2in}@{}}
			\toprule
			\multicolumn{1}{c}{Yea - Nay} & \multicolumn{1}{c}{Outcome} & \multicolumn{1}{c}{Purpose}  \\ \midrule
			97 - 0  & Agreed to & A resolution honoring the victims and heroes of the shooting\ldots in Tucson, Arizona. \\ \hline
			96 - 1  & Agreed to & \ldots to provide penalties for aiming laser pointers at airplanes\ldots \\ \hline 
			100 - 0 & Confirmed & Confirmation Robert S. Mueller, III, of California, to be Director of the Federal Bureau of Investigation \\ \hline 
			0 - 97 & Rejected & \ldots budget request for the United States Government for fiscal year 2012, and\ldots budgetary levels for fiscal years 2013 through 2021. \\ \bottomrule
		\end{tabular}
	\end{table}
}
Since a significant portion of votes is nearly unanimous and/or procedural as according to senate rules, as opposed to actually debated and/or legislative and thus truly informative, it is unclear that modeling the effect that temporally neighboring votes have on each other is more meaningful than recovering the reciprocal relations (at least using the frameworks presented in this chapter). For some examples of such unanimous or procedural votes, see table~\ref{tab:votes}~\cite{noauthor_u.s._2011}. Thus, we apply the undirected graph estimation model from equation~\eqref{eq:extemp_reg} with the low-rank component included to track the network of relations among the senate.

We use a slightly different sparsity regularization to account for the symmetric structure of the adjacency matrix to be estimated,
\begin{align}
\label{eq:reg_symm}
\textrm{h}(\Acal)= \sum\limits_{k}\sum\limits_{i<j} \left\|([\A_k]_{ij} \; [\A_k]_{ji})\right\|_2.
\end{align}

\begin{figure}[!h]
	\includegraphics[width=0.95\columnwidth]{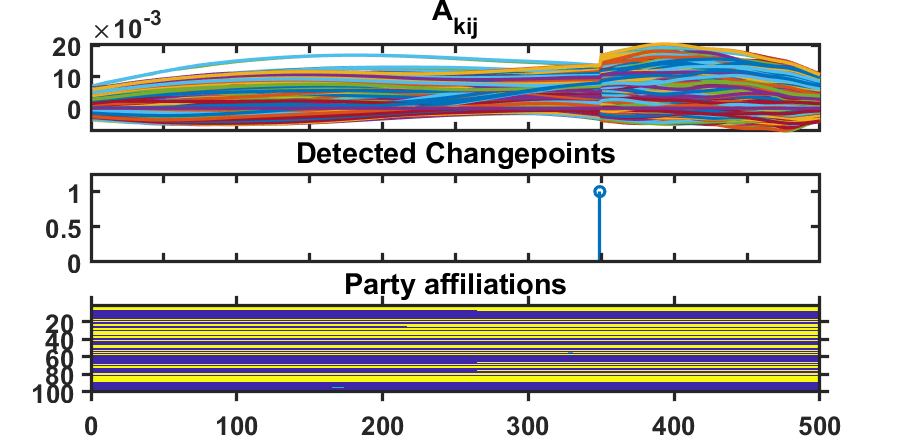}
	\caption{Top: time series of entries of estimated graph adjacency. Each color is one time element of the matrix. Middle: Changepoints detected by algorithm. Bottom: Actual party affiliations for each seat, yellow for Republican, blue for Democrat.}
	\label{fig:pna:sens_timeseries_chgpt} 	
\end{figure}
We see at the top of figure~\ref{fig:pna:sens_timeseries_chgpt} the $N^2=10^4$ individual elements of the estimated adjacency matrix, one color for each $ij$ entry plotted as a function of time $k$. In the middle, we see that the algorithm detected a changepoint at timepoint 349. In the bottom of the figure, we see evidence that there was actually the start of a new session (transition from 111-112) at timepoint 264 (Jan. 2011) in which 5 new senators were elected of the opposite political party to the previous senator they replaced (Republicans replacing Democrats). This happened as the so-called ``Tea Party,'' which started as a vocal grassroots movement and was active in organizing national demonstrations, had gained major traction in the US Republican Party~\cite{karpowitz_tea_2011}. The estimate of the changepoint is not in fact wrong; since the available data is the voting record in the senate, what this analysis shows is that the impact of the election is felt not immediately after but after initial pro-forma votes till the senate settles down to its real business  of drafting and voting on impactful legislation. The lag in detection corresponds to the extended segment of largely uninformative almost unanimous votes in the beginning of the new session (see figure~\ref{fig:votes} where, between columns 264 and 349, columns are mostly of the same color, either yellow or blue, with few dissents) so that the change only really manifests itself after this intial stage with the senate settling into the new session.  

\begin{figure}[!ht]
	\includegraphics[width=0.95\columnwidth]{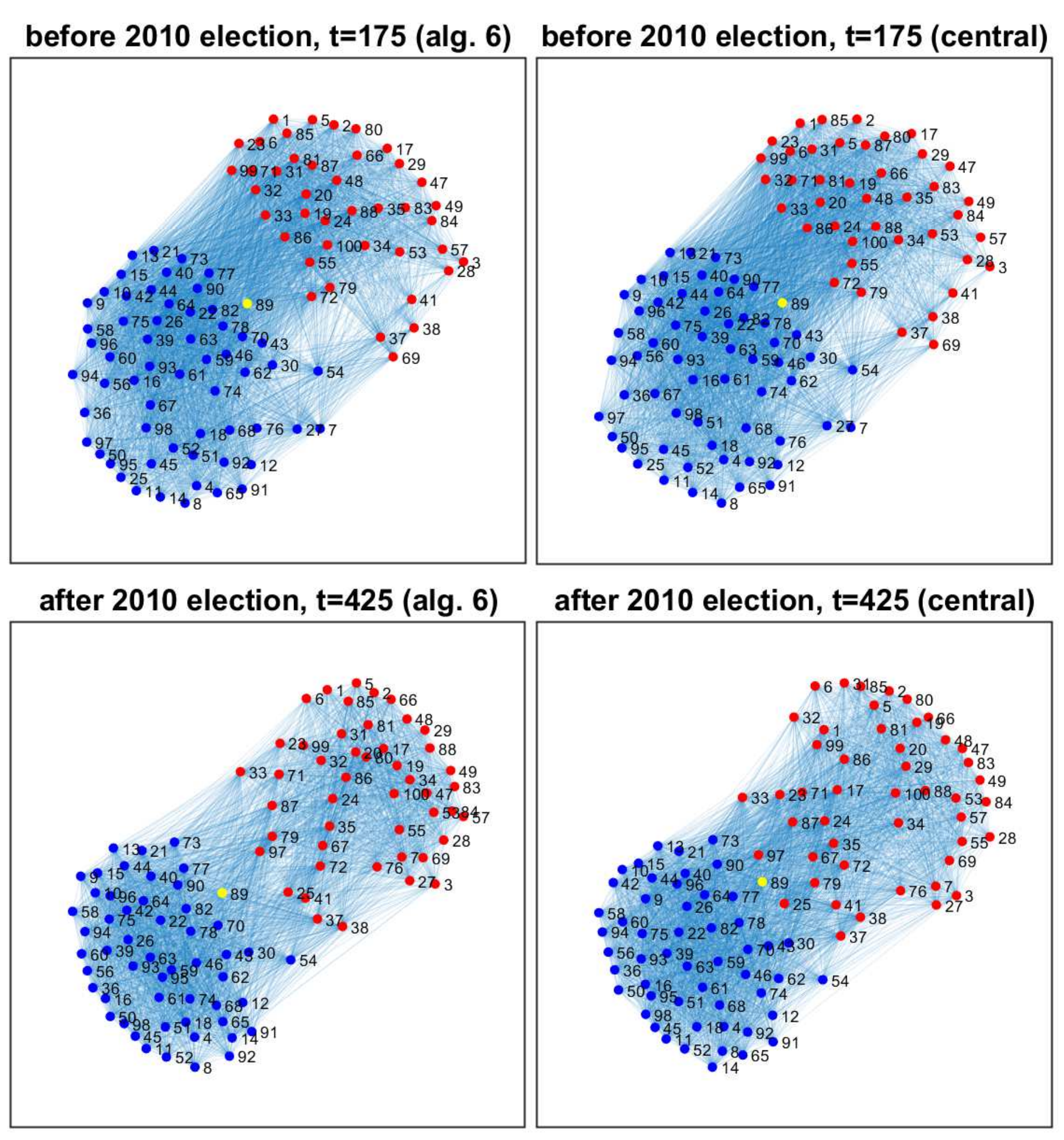}
	\caption{Top: Estimated adjacency graph from middle of first session; Bottom: Estimated adjacency graph from middle of second session; Left: Our method; Right: Central estimator}
	\label{fig:pna:sens_mid_session} 	
\end{figure}

\begin{figure}[!h]
	\includegraphics[width=0.95\columnwidth]{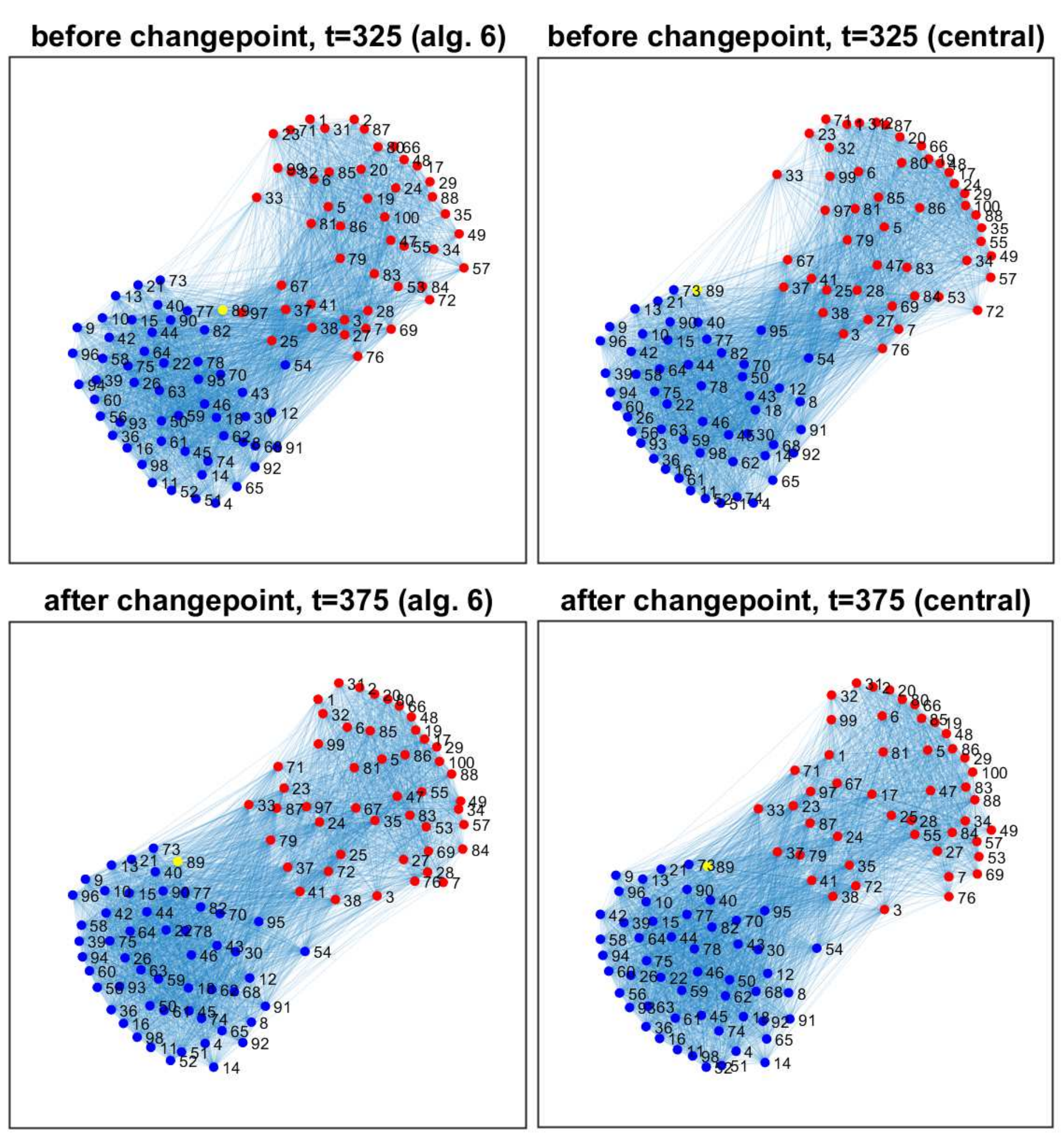}
	\caption{Top: Estimated adjacency graph from the first session near the detected changepoint; Bottom: Estimated adjacency graph from the second session near the detected changepoint; Left: Our method; Right: Central estimator}
	\label{fig:pna:sens_chgpt}
\end{figure}

In figures~\ref{fig:pna:sens_mid_session} and~\ref{fig:pna:sens_chgpt}, we compare estimated networks using our method and the central kernel estimate from $k=175$ and $k=425$, corresponding to two points respectively before and after the start of the new session as well as the detected changepoint. The top layouts are achieved by using a planar force-directed embedding based on the edges estimated in $\what{\A}_k$ (in particular, we use the Fruchterman-Reingold~\cite{fruchterman_graph_1991} layout as implemented in MATLAB\textregistered~version 2018a). Intuitively, shorter average path lengths between nodes correspond to larger forces and lower distances in the planar embedding. The bottom layouts are computed using the top layouts as initialization points. Neighboring points vote more similarly, signifying ideological closeness. We additionally label seats according to the ground truth party of the senator occupying them at that time, with Republicans in red, Democrats in blue, and the Independent in yellow.

In figure~\ref{fig:pna:sens_mid_session}, we can see that the graph clearly shows the polarization between the two parties. Interestingly, the Independent is shown as ideologically closer to the democrats but on the closest edge to the Republicans. Indeed, the Independent is Senator Sanders, whom we know to hold views that are espoused by both parties, but in 2016 ultimately ran in the Democratic primary race to try to become the party's presidential candidate. In addition, we point out two seats, $7$ (Arkansas) and $27$ (Indiana) who were Democratic prior to the election, but switched to Republican afterwards. These senators were on the boundary of the Democratic cluster to begin with, which is consistent with the fact that both states historically lean Republican as measured by the previous $8$ presidential elections (2008 was in fact an exception for Indiana, while 1992 and 1996 were exceptions for Arkansas as the Democratic candidate was from Arkansas)~\cite{noauthor_arkansas_2018,noauthor_indiana_2018}. Also, we note that the shapes of the clusters change, with the Democratic clique flattening while the Republican clique stretches. We must be careful in drawing any conclusions from this phenomenon, but this could weakly support (i.e., not provide any evidence to disprove) the conclusions drawn by political researchers claiming that official recognition of the ``Tea Party Movement'' and its unity resulted in the Republican Party moving ideologically further away from the previous center of the political space~\cite{williamson_tea_2011,gervais_reading_2012}. Finally, we see that the two methods estimate very similar graphs, showing that our method performs a similar estimation to the smooth central kernel estimator in regimes away from changepoints. 

In figure~\ref{fig:pna:sens_chgpt}, we visualize graphs estimated on either side of the detected changepoint at $k=349$, this time much closer to the changepoint. We see that the central kernel shows less change between the two timepoints, while there is a noticeable difference across the changepoint in our method. In the top row, there is also some movement relative to the timepoints visualized in figure~\ref{fig:pna:sens_mid_session}, possibly corresponding to posturing ahead of the election. Continuing our previous interpretations, this could be the existing Republicans trying to maintain their seats over ``Tea Party'' primary challengers~\cite{williamson_tea_2011}. We clarify one subtlety on this theory: though the visualized timepoint is actually \textit{after} the beginning of the new session, the kernel bandwidth is such that the votes from \textit{before} in the old session still influence the estimated graph in both methods, as the visualized timepoint still occurs before the \textit{detected} changepoint. 
\begin{figure}[!h]
	\includegraphics[width=0.95\columnwidth]{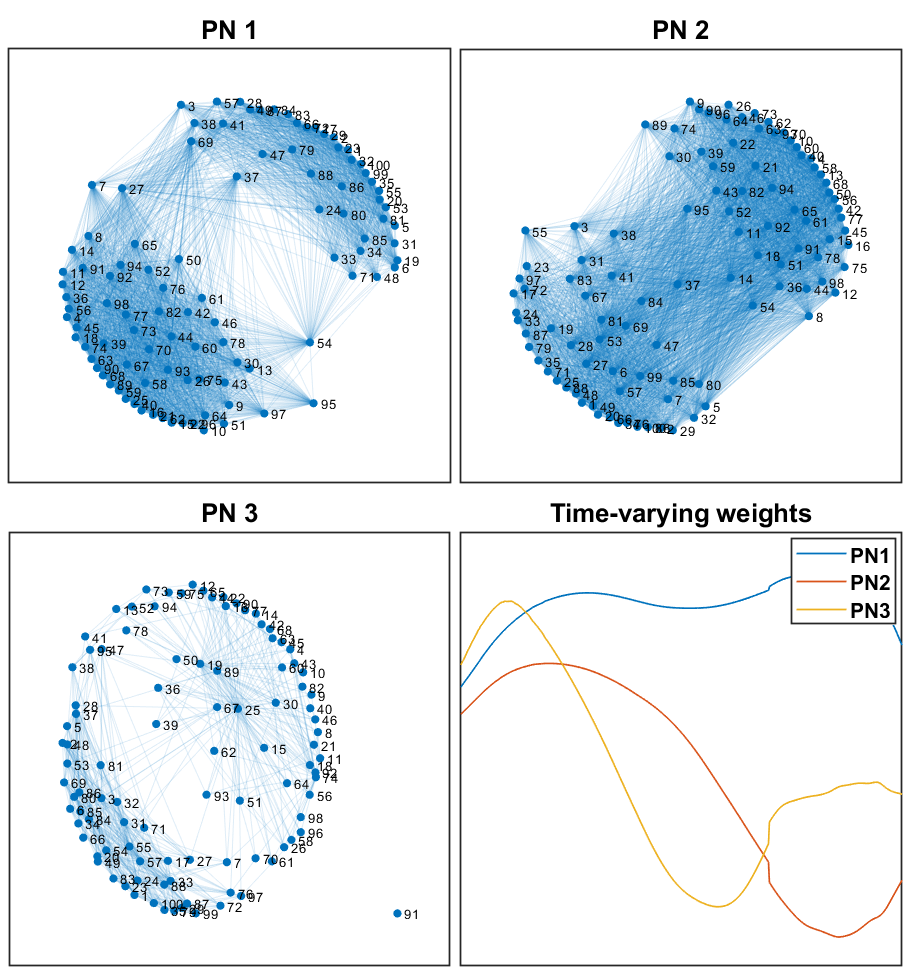}
	\caption{Principal Networks of the senate voting record; Bottom-right: relative time-varying weights for each network}
	\label{fig:pna:sens_pn}
\end{figure}

Finally, in figure~\ref{fig:pna:sens_pn} we visualize the result of PNA applied to this dataset, again using a force-directed layout. This time, note that we do not label the nodes by color since these networks are now across all time, so several seats could be either red or blue depending on the time. We find that the first two Principal Networks (PN) capture the main overall polarized structure of the two parties. PN 3, on the other hand, seems to highlight the most salient and influential seats in the senate. In this case, we see visually that the time series for PN 3 has the largest time derivatives, so that the network corresponds to the edges that had the largest changes throughout the dataset. Again, Independent senator Sanders (89) is seen towards the interior of the network, indicating the existence of many edges with many seats with all manner of ideologies. We see that seat 25 is also towards the center of this network. Interestingly, this is Democratic senator Burris from Illinois, who was specially appointed by the governor as a result of then-senator Obama's resignation after his election to the presidency in 2008, who retired prior to the beginning of the new session to be replaced by Republican senator Kirk. This change in ideologies occurring before the single detected changepoint was clearly not large enough in magnitude to result in another changepoint, but rather it led to the seat seeming to share ideologies across both parties and a strong presence in PN 3 as a seat with many large enough edges changing smoothly.

\subsection{Gene Interaction Networks}
We examine genetic expression data for \textit{Drosophila Melanogaster}~\cite{arbeitman_gene_2002} that covers the 4 stages of its life cycle: egg, larva, pupa, adult. The expresion dataset collects binary values for $N'=4028$ genes at $K=66$ time points. The egg stage corresponds to time points $1 \le k \le 30$, the larva to time points $31 \le k \le 40$, the pupa to time points $41 \le k \le 59$, and the adult to time points $60 \le k \le 66$. Of course, trying to estimate a network on so many genes is ill-posed, even when trying to impose high levels of sparsity. Thus, as with previous attempts at network analysis on similar gene expression data~\cite{kolar_graph_2014}, we select a subsample of developmental genes. We analyze $N=25$ random genes. In the context of this dataset, for ease of presentation to a non-biologist audience, we overload the term ``gene'' to refer to either genes or the proteins they encode, and label them interchangeably in figures.

We again use the symmetric regularization in equation~\eqref{eq:reg_symm} to estimate the interaction networks or eigennetworks. This is because by subsampling we have less ability to determine causality via directed networks, even when including the low-rank network components corresponding to latent variables. However, we see that we are still able to glean meaningful information from the data. 

\begin{figure}[htb]
    \begin{subfigure}[t]{0.5\columnwidth}
	    \includegraphics[width=0.95\columnwidth]{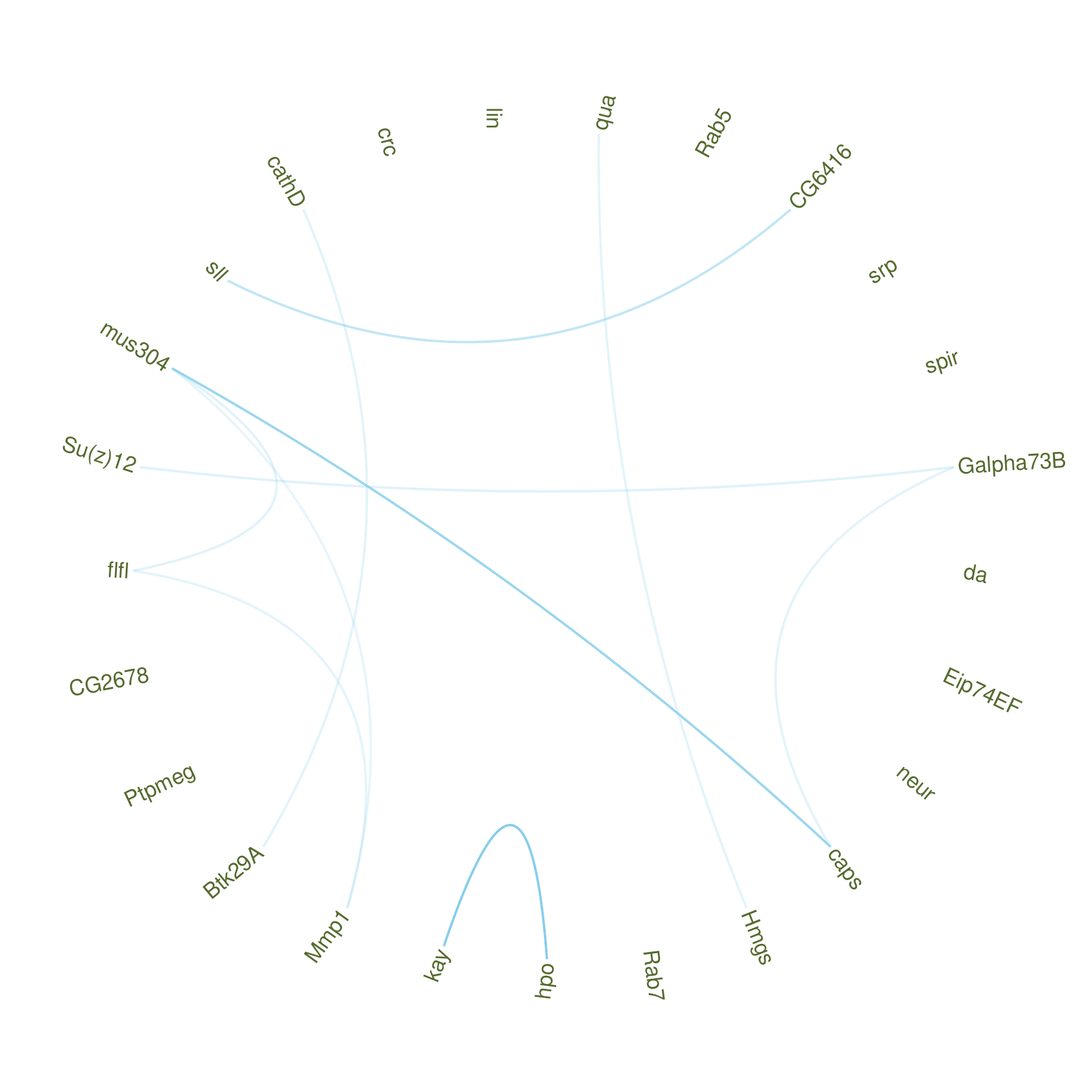}
	    \caption{Net 1}
	    \label{fig:pna:gene_nets:1}
    \end{subfigure}%
    ~
    \begin{subfigure}[t]{0.5\columnwidth}
	    \includegraphics[width=0.95\columnwidth]{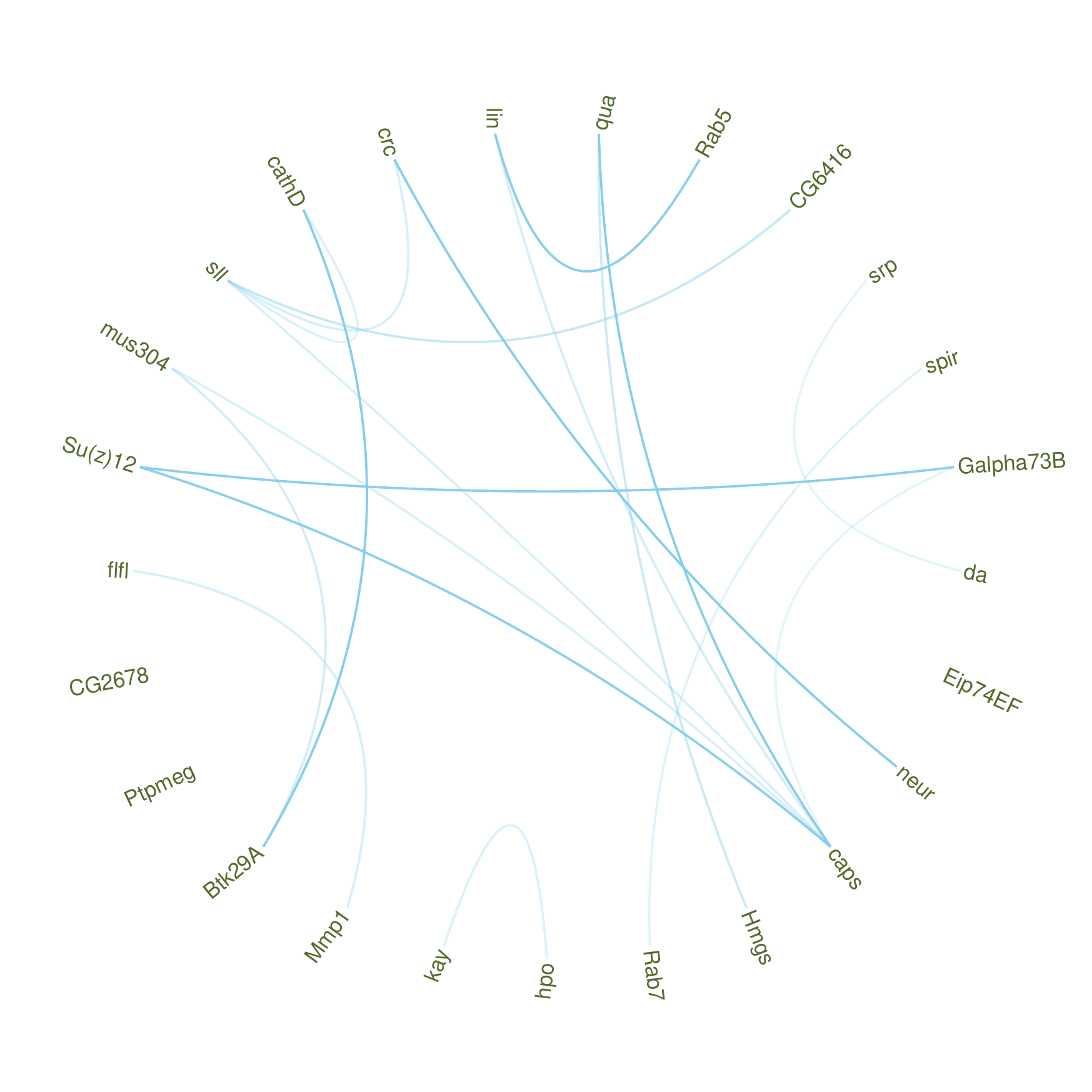}
	    \caption{Net 2}
	    \label{fig:pna:gene_nets:2}
    \end{subfigure}
    
    \begin{subfigure}[t]{0.5\columnwidth}
	    \includegraphics[width=0.95\columnwidth]{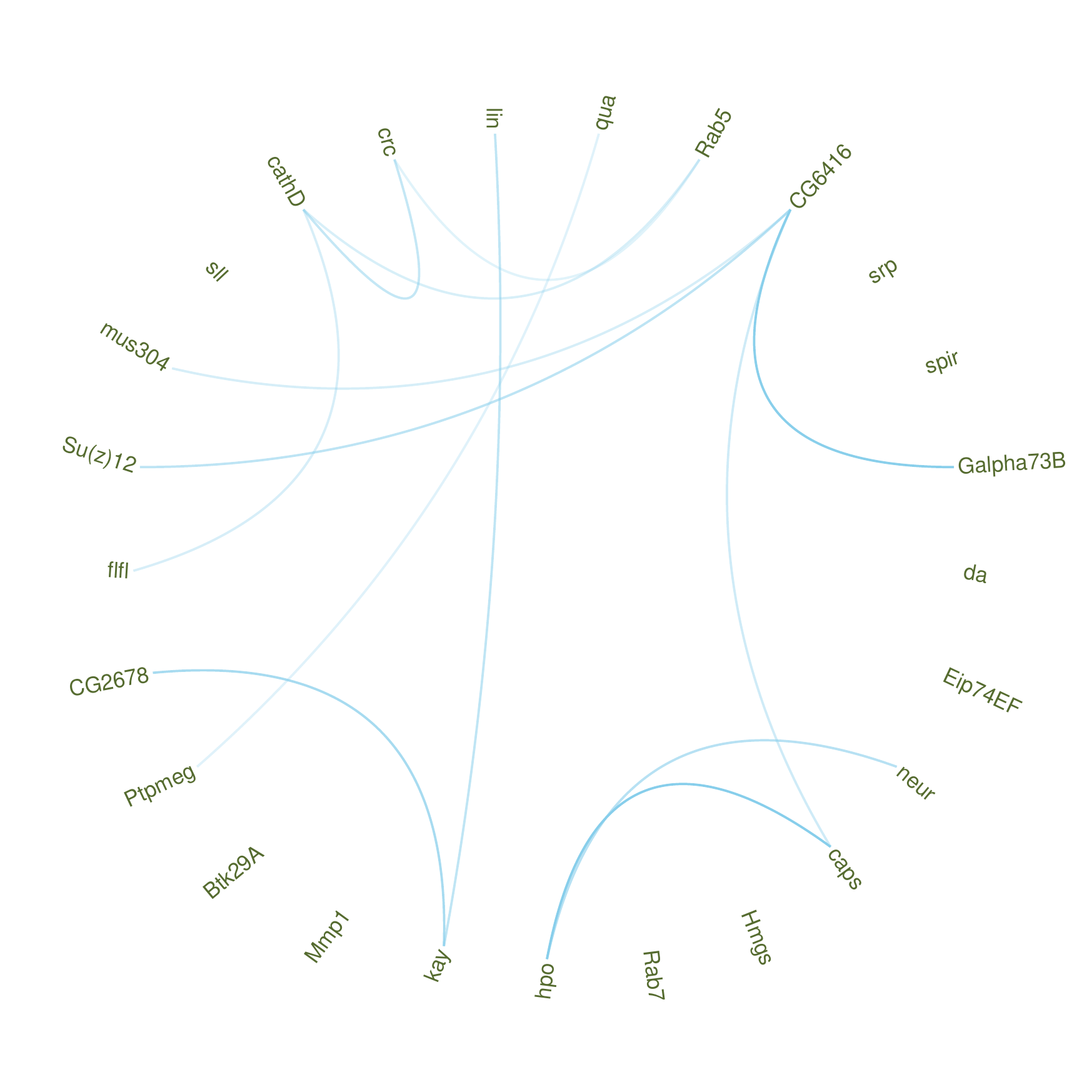}
	    \caption{Net 3}
	    \label{fig:pna:gene_nets:3}
    \end{subfigure}%
    ~
    \begin{subfigure}[t]{0.5\columnwidth}
	    \includegraphics[width=0.95\columnwidth]{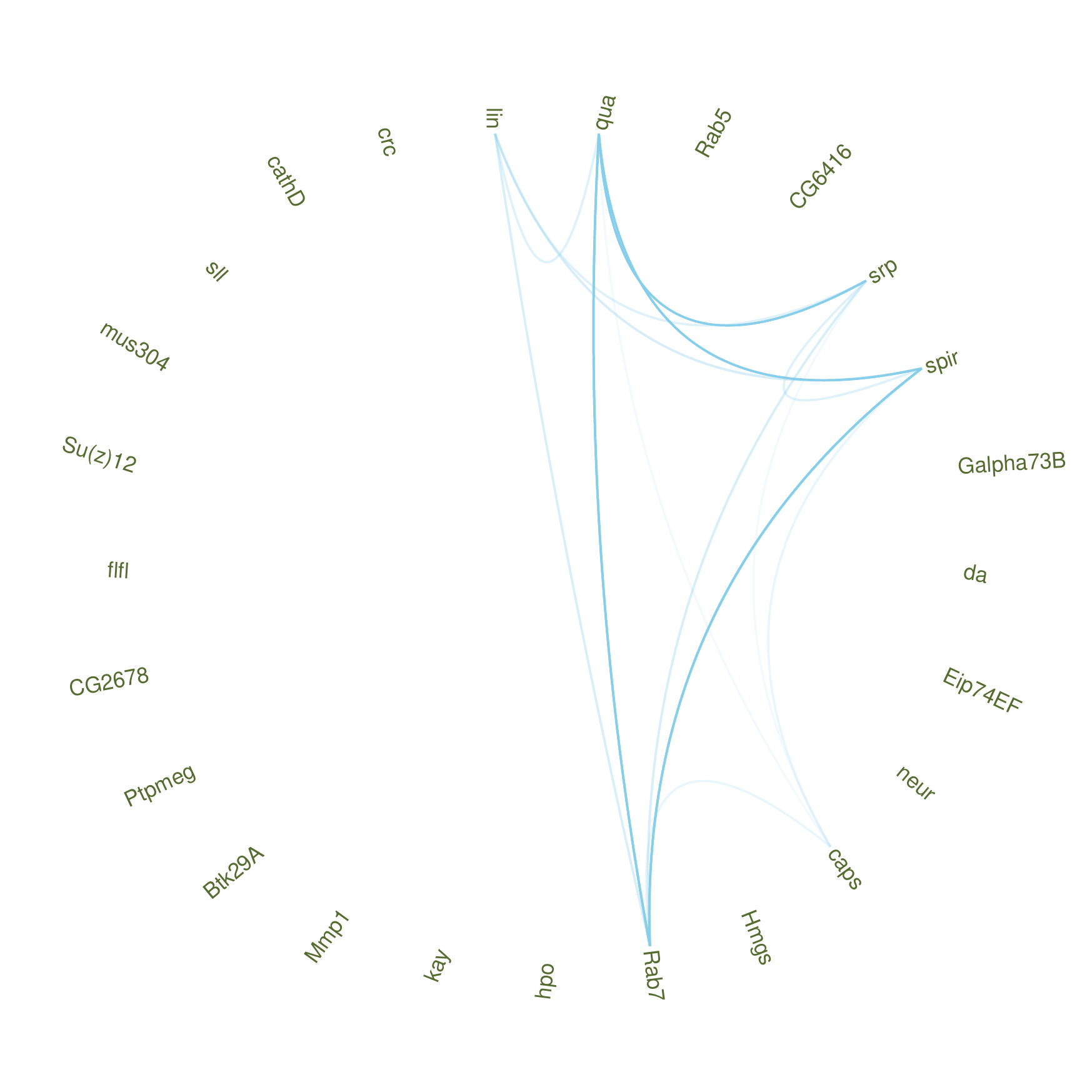}
	    \caption{Net 4}
	    \label{fig:pna:gene_nets:4}
    \end{subfigure}
	\caption{The weights for the EigenNetworks estimated from gene expression data}
	\label{fig:pna:gene_nets}
\end{figure}

\begin{figure}[htb]
	\includegraphics[width=0.95\columnwidth]{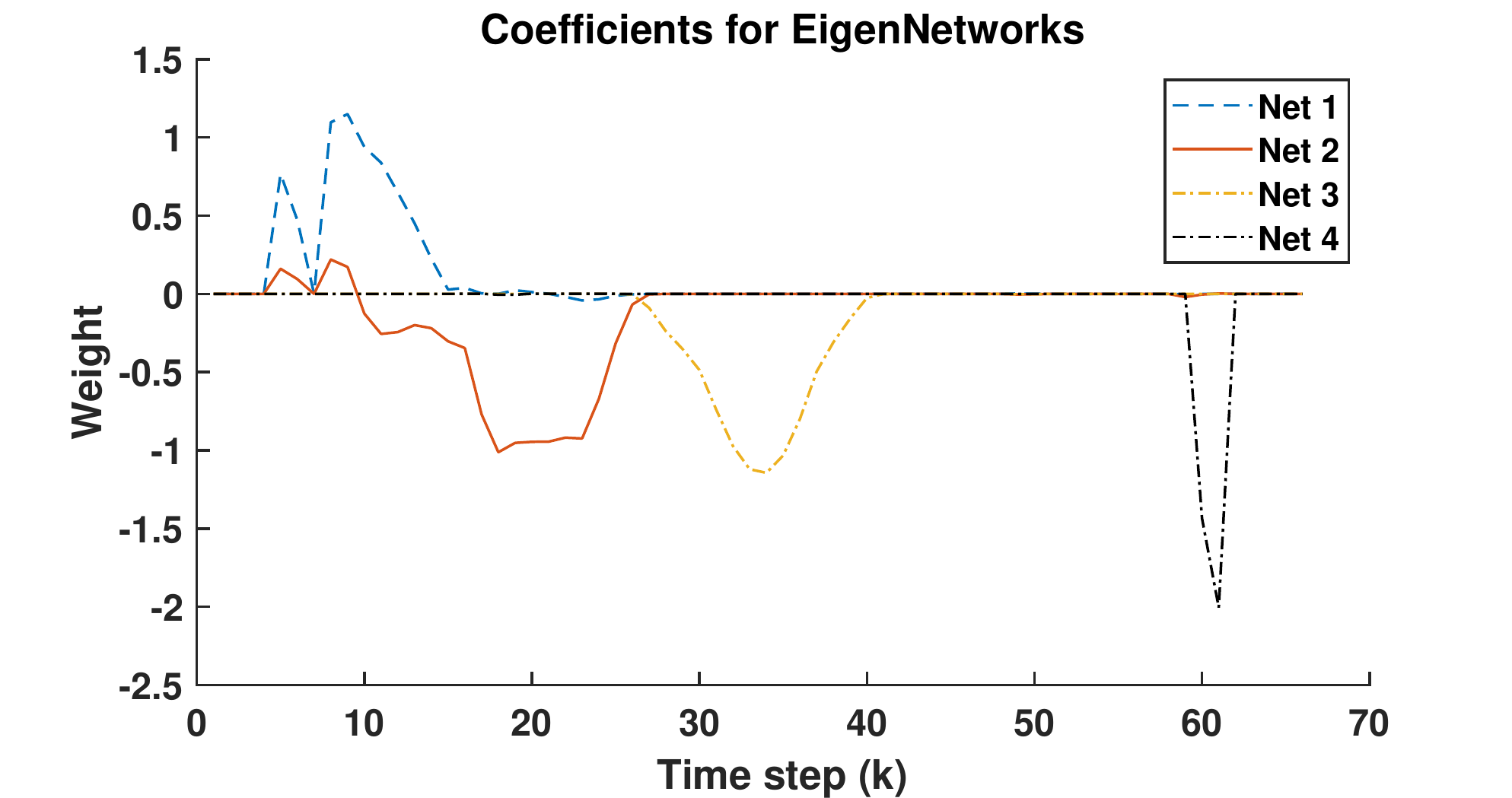}
	\caption{The weights for the eigennetworks estimated from gene expression data}
	\label{fig:pna:gene_coeffs}
\end{figure}

Figure~\ref{fig:pna:gene_nets} shows the eigennetworks estimated from the data displayed using a hierarchical edge bundling~\cite{holten_hierarchical_2006} layout in R~\cite{r_core_team_r:_2014} and figure~\ref{fig:pna:gene_coeffs} shows their corresponding weights through time.

In figure~\ref{fig:pna:gene_nets:1} in the first eigennetwork, we point out the detection of a connection detected between the ``kay'' and ``hpo'' genes. These both interact directly with gene ``sd'', which we do not have in this subsample but is part of a signalling pathway that determines organ size during early development~\cite{rohith_scalloped_2017}. In figure~\ref{fig:pna:gene_nets:2}, the second network shows the genes that have high activity during larval development activated. In particular, it is worth noting that the ``caps'' gene most apparently central in this network has been specifically implicated in pre-pupal wing development, which is one of the major aspects of \textit{Drosophila} metamorphosis and supported by other active processes in the pre-pupal stages~\cite{milan_lrr_2001}. In figure~\ref{fig:pna:gene_nets:3}, the third network shows that ``caps`` is still active, but now interacting with ``hpo'', which also has a role in photoreceptor neuronal (eye) development~\cite{nolo_bantam_2006}. Interestingly, ``caps'' has a secondary role after pre-pupal wing development, which is precisely pupal retinal development~\cite{shinza-kameda_regulation_2006}. We further note that ``hpo'' and ``sd'' also play a role in cancer suppression later on while ``kay'' is not known to, and we see both ``hpo'' and ``kay'' do stay active but now in separate subnetworks rather than directly connected as in the first EigenNetwork. Finally, in figure~\ref{fig:pna:gene_nets:4} we get to see the few genes active during adulthood. For example ``srp'' is responsible for glucose metabolism, and ``spir'' is implicated in the formation of new eggs (oogenesis)~\cite{quinlan_drosophila_2005}, and the development of the new embryo from the new egg depends on proper energy storage by the mother~\cite{fraga_glycogen_2013}. 

We note that changepoints can be detected detected at time points 7, 39, and 58. The 39 and 58 are consistent with the transitions from larva to pupa and from pupa to adult, while the changepoint at 7, though not one of the four major life stages, is due to the fact that these first 7 time steps actually represent the first few ``highly dynamic'' hours between fertilization and gastrulation; the genes that initially express are maternal, followed by some genes that express only after fertilization actually due to the new zygote~\cite{arbeitman_gene_2002}. The transition from egg to larva at 30 is not detected, but we see the evidence of the change still present in the time series of coefficients. In fact, we can see this as evidence that the change in gene expression during the transition from egg to larva is actually relatively smooth rather than abrupt at this time resolution, which is again consistent with previous gene expression studies~\cite{arbeitman_gene_2002}. We also note that this subsample unfortunately does not contain any gene networks that are significantly active during the pupal stage. Still, the time series of network weights that the algorithm detects corresponds to distinct regimes that in fact line up well with the other known developmental stages. 

\section{Conclusion}
\label{sec:conc}
We have presented a framework for learning the principal networks or eigennetworks--a small set of fundamental graphs--and eigenfeatures that underlie a system described by smooth graphs with potential changepoints. This framework includes a time-varying graph estimation method that is easily parallelized, as a middle ground to existing methods for estimating time-varying graphs. The framework additionally uses a provably convergent matrix factorization to find the fundamental or eigengraphs. We demonstrate the use of this framework on simulated data as well as
\begin{inparaenum}[1)]
\item real US senate voting records for identifying interesting entities within the network and the times at which salient events occurred; and
    \item genetic expression data for the \textit{Drosophila Melanogaster} as it goes through its four stages of metamorphosis. 
        \end{inparaenum}

An interesting direction for future work includes developing additional guarantees for each step of the approach and thus observing the propagation of error through the two steps. Additionally, the application of this method to other areas of science would be exciting, especially in budding areas of neuroscience. A final area of interest is in combining the frameworks for learning graph time series and for performing matrix factorization into a single problem in a numerically stable algorithm, again with performance guarantees.

\ifCLASSOPTIONcaptionsoff
\newpage
\fi

\bibliographystyle{IEEEtran}
%\bibliography{jmei,PostPhD,additional}

\end{document}